\documentclass[10pt,twocolumn,letterpaper]{article}
\usepackage[pagenumbers]{cvpr} 

\usepackage{xspace}
\usepackage{xcolor} 
\definecolor{myblue}{rgb}{0.88,0.98,1}
\definecolor{mygreen}{rgb}{0.92, 1.0, 0.92}
\definecolor{midgreen}{HTML}{69c5a3}
\definecolor{midblue}{HTML}{69a3f1}
\definecolor{darkgreen}{HTML}{146038}
\definecolor{darkblue}{HTML}{143b59}
\usepackage{multirow, booktabs, graphicx, colortbl, arydshln, tikz}
\definecolor{cvprblue}{rgb}{0.21,0.49,0.74}
\def\etc{\emph{etc}\onedot} 
 
\def\eg{\emph{e.g}\onedot} 

\newcommand{\ours}[1]{EECA}
\usepackage{amsmath}
\usepackage[pagebackref,breaklinks,colorlinks,allcolors=cvprblue]{hyperref}
\usepackage{makecell}
\title{\textsl{Beyond Sight}: Towards Cognitive Alignment in LVLM via Enriched Visual Knowledge}

\author{
Yaqi Zhao$^1$\footnotemark[1], Yuanyang Yin$^2$\footnotemark[1], Lin Li$^1$, Mingan Lin$^3$, Victor Shea-Jay Huang$^1$\\
Siwei Chen$^1$, Weipeng Chen$^3$, Baoqun Yin$^2$, Zenan Zhou$^3$\footnotemark[2], Wentao Zhang$^1$\footnotemark[2]\\
{$^1$Peking University\hspace{0.5cm}}
{$^2$University of Science and Technology of China\hspace{0.5cm}}
{$^3$Baichuan Inc.\hspace{0.5cm}}\\
}

\begin{document}
\maketitle
\renewcommand{\thefootnote}{\fnsymbol{footnote}}
\footnotetext[1]{Equal contribution.}
\footnotetext[2]{Corresponding Author.}
\begin{abstract}
Does seeing always mean knowing? Large Vision-Language Models (LVLMs) integrate separately pre-trained vision and language components, often using CLIP-ViT as vision backbone.  However, these models frequently encounter a core issue of ``cognitive misalignment" between the vision encoder (VE) and the large language model (LLM). Specifically, the VE's representation of visual information may not fully align with LLM's cognitive framework, leading to a mismatch where visual features exceed the language model’s interpretive range.
To address this, we investigate how variations in VE representations influence LVLM comprehension, especially when the LLM faces VE-Unknown data—images whose ambiguous visual representations challenge the VE’s interpretive precision. Accordingly, we construct a multi-granularity landmark dataset and systematically examine the impact of VE-Known and VE-Unknown data on interpretive abilities. Our results show that VE-Unknown data limits LVLM’s capacity for accurate understanding, while VE-Known data, rich in distinctive features, helps reduce cognitive misalignment.
Building on these insights, we propose Entity-Enhanced Cognitive Alignment (EECA), a method that employs multi-granularity supervision to generate visually enriched, well-aligned tokens that not only integrate within the LLM's embedding space but also align with the LLM's cognitive framework. This alignment markedly enhances LVLM performance in landmark recognition. Our findings underscore the challenges posed by VE-Unknown data and highlight the essential role of cognitive alignment in advancing multimodal systems.
\end{abstract}    
\section{Introduction}
\label{sec:intro}
\begin{figure*}[t!]
    \centering
    \includegraphics[width=\textwidth]{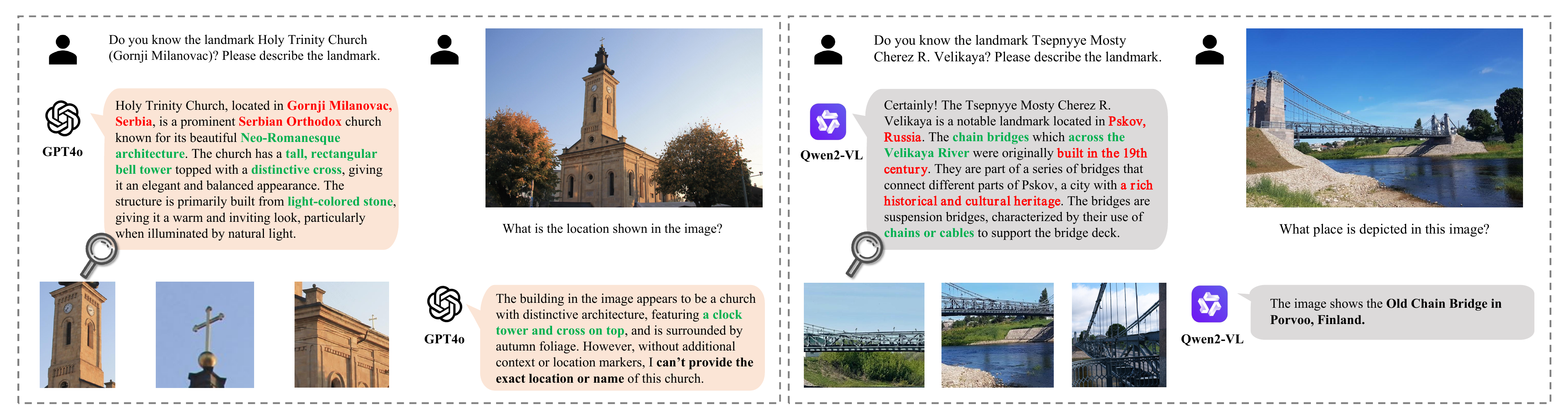}
    \caption{Instances of \textit{Cognitive misalignment} are systematically identified, even in advanced models like GPT-4o and Qwen2-VL. Although the image closely aligns with the description generated from the text-only prompt, both models fail to recognize the landmark when presented with the image. Text highlighted in \textbf{\textcolor{OliveGreen}{green}} emphasizes details that are particularly relevant to the image.}
     \vspace{-2mm}
    \label{fig:qa_example}
\end{figure*}
Large Vision Language Models (LVLMs)~\citep{gpt4o, Bard, liu2023improved, dai2023instructblip,li2024mini,blip2,laurenccon2024matters,lu2024deepseek,xue2024xgen,zhang2024internlm} have recently achieved significant advancements. By mapping visual inputs into the embedding space of large language models, LVLMs harness the powerful interpretative capabilities of language models~\citep{openai2023gpt4, llama3modelcard, qwen2.5} to address complex tasks like visual question answering, grounding, counting, \etc. 

Despite the impressive advancements in LVLMs, these models still struggle with fundamental recognition tasks. As illustrated in~\Cref{fig:qa_example}, even state-of-the-art models like GPT-4o and Qwen2-VL fail to recognize iconic landmarks from images, although they can describe these landmarks accurately when prompted with text alone. This raises critical questions: {\em Why do these challenges persist? To what extent do these models truly understand what they perceive?} Drawing inspiration from~\citep{gekhman2024does}, we attribute this issue to a broader problem that we call \textbf{\em cognitive misalignment}—a fundamental disconnect between the representations generated by the vision encoder (VE) and the interpretive framework of the large language model (LLM).

Currently, most LVLMs~\citep{dai2023instructblip, liu2023improved, bai2023qwen} integrate separately \emph{pretrained} vision~\citep{radford2021learningtransferablevisualmodels, sun2023eva, zhai2023sigmoid} and language models~\citep{llama3modelcard, peng2023instruction, qwen2.5} through different adapters~\citep{flamingo, blip2, liu2023improved} to achieve multimodal comprehension.
However, this simple connection often results in fundamental misalignment, as recent studies suggest~\citep{yin2024sea, tong2024eyes, wei2025vary}, which limits the potential of LVLMs.
In this paper, we start by examining the cognitive framework of the vision encoder (VE) in LVLMs (\Cref{sec:sight2insight}). To systematically investigate this, we define evaluation metrics based on CLIP’s training paradigm to assess VE knowledge. 
Using these metrics, we categorize data into VE-Known and VE-Unknown. Our study empirically demonstrates the following insights:
\begin{itemize} 
    \item \textbf{Enhanced alignment with VE-Known data}: VE-Known data, comprising images with rich and distinctive features, provides a strong, discriminative foundation that enhances the utilization of visual knowledge in LVLMs. This data enables smoother alignment between the vision encoder (VE) and the language model (LLM), allowing for more effective cognitive integration.
    \item \textbf{Challenges with VE-Unknown data}: In contrast, VE-Unknown data, characterized by weak alignment with textual labels, poses significant alignment challenges. These representations hinder the VE’s ability to align with the LLM, leading to reduced interpretability and performance degradation on downstream tasks.
    \item \textbf{Importance of data quality over quantity}: The contrast between VE-Known and VE-Unknown data underscores that data quality has a greater impact than sheer volume. VE-Unknown samples tend to hinder cognitive integration, limiting the LVLM’s ability to effectively process complex visual inputs. This finding highlights the need to prioritize high-quality, VE-Known data to maximize LVLM performance.
\end{itemize}

Having identified the cognitive framework of the vision encoder in LVLM, we shift our focus toward enriching the visual knowledge in a way that aligns with the LLM’s cognitive framework, effectively ``opening the eyes'' of the LLM and achieving cognitive alignment (\Cref{sec:method}). To this end, we construct \textbf{M}ulti-\textbf{g}ranularity \textbf{L}andmark \textbf{D}ataset (MGLD) that ensures consistency between the visual input and the language model's output. Building on this, we propose the Entity-Enhanced Cognitive Alignment (\ours{}), a structured approach that enhances the richness and discriminative power of visual tokens through supervised multi-granularity learning. 
Through this multi-faceted approach, EECA not only minimizes information loss in the adapter’s transformation but also enables the transformed visual tokens to mimic VE-Known representations. 

Our contributions are as follows. First, we provide an in-depth analysis of the impact of VE representations on the LLM’s understanding, shedding light on the challenges of cognitive alignment. Building on these insights, we propose \ours{}, a novel method that introduces discriminative features from the LLM’s recognition space into visual tokens through supervised learning, enhancing their effectiveness for cognitive alignment. Finally, extensive experiments and analyses show that our method significantly enhances landmark recognition performance, with notable improvements observed on both VE-Known and VE-Unknown data, consistently outperforming baseline models.

\section{Preliminary}\label{sec:pre}
\paragraph{Notation.}
In typical Large Vision Language Models (LVLMs)~\citep{xue2024xgen,dai2023instructblip,liu2024sphinx,lu2024deepseek,dong2024internlm,laurenccon2024matters,liu2023improved,liu2023visual,ye2024mplug,wang2024qwen2}, an \textbf{adapter} is used to connect VE and LLM seamlessly.
This adapter, denoted by \( g_{\theta} \), can be a simple linear layer or a more complex attention module. For clarity, we define the features output by the vision encoder as \textit{visual patches} and the features after passing through the adapter as \textit{visual tokens}.

In our framework, the vision encoder \( f_v \) and text encoder \( f_t \) are from CLIP’s dual modules. The vision encoder \( f_v \) processes an image \( I_i \) to produce \( V_i = f_v(I_i) \in \mathbb{R}^{P \times d} \), where \( d \) is the feature dimension and \( P \) is the number of visual patches. The text encoder \( f_t \) maps text \( T_i \) to \( t_i = f_t(T_i) \in \mathbb{R}^d \).
During LVLM training, the LLM input is constructed as follows:
\begin{equation}
    X_v = g_{\theta}(f_v(I_i)), \quad X_t = \phi(T_i),
\end{equation}
where \( g_{\theta} \) transforms the vision encoder’s output to visual tokens \( X_v \in \mathbb{R}^{N_v \times C} \), and \( \phi \) maps text \( T_i \) to text tokens \( X_t \in \mathbb{R}^{N_t \times C} \) in the LLM’s embedding space. The visual tokens \( X_v \) and text tokens \( X_t \) are then concatenated and fed into the LLM to generate a response.
\begin{figure*}[htbp]
    \centering
    \includegraphics[width=\textwidth]{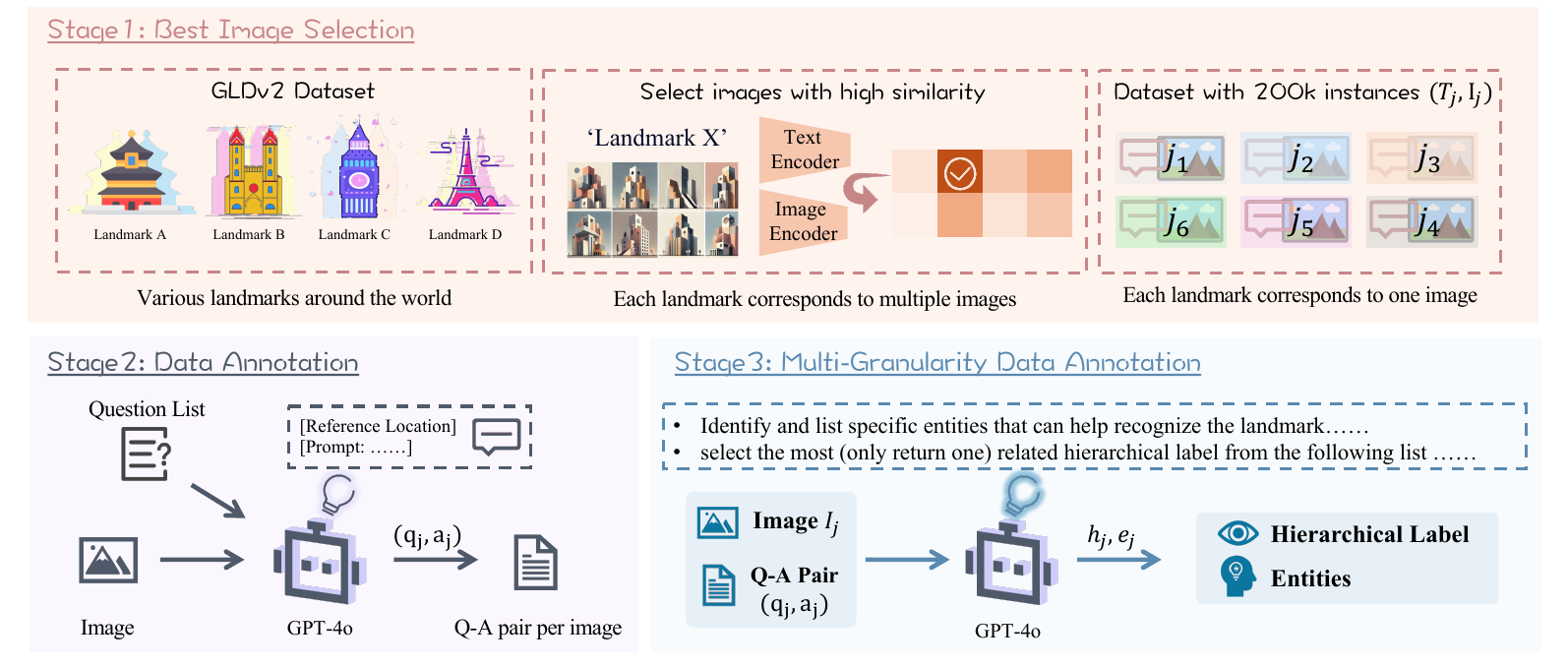}
    \caption{Illustration of the dataset construction process for the Multi-granularity Landmark Dataset (MGLD), showing three stages: best image selection using CLIP similarity, data annotation with Q-A pairs, and multi-granularity data annotation.}
     \vspace{-2mm}
    \label{fig:pipeline}
\end{figure*}

\paragraph{Cognitive misalignment: issues and challenges.}
We identified a critical issue in LVLMs, illustrated through the example in ~\Cref{fig:qa_example}, which reveals a phenomenon we term \textit{cognitive misalignment}. This issue arises from discrepancies between Vision Encoder's representations and the language model's cognitive framework. For instance, when asked, ``Do you know the landmark Holy Trinity Church (Gornji Milanovac)? Please describe the landmark,'' the model provides an accurate description based on its textual knowledge. However, when shown an image that corresponds to this description and asked, ``What is the location shown in the image?'', it fails to recognize it. This misalignment highlights a core challenge: although the language model can describe landmarks accurately in text, the VE’s visual outputs do not align closely enough with the language model’s cognitive framework to enable reliable recognition from images. This insight drives our exploration of different types of visual knowledge—\textbf{\textit{VE-Known}} and \textbf{\textit{VE-Unknown}}—which we elaborate on in later sections.
\section{From Sight to Insight} \label{sec:sight2insight}
In this section, we explore the cognitive alignment between the VE and the LLM within LVLMs. By constructing a fine-tuned dataset and evaluating visual knowledge across different levels, we aim to systematically analyze the quality of visual outputs and their impact on achieving cognitive alignment with the LLM's understanding.
\subsection{Study setup}\label{subsec:setup}
\paragraph{Dataset construction.} 
To explore cognitive alignment between Vision Encoders (VE) and large language models (LLMs) in LVLMs, we use an entity-related dataset where both the visual and language models have their own distinct representations and understandings. This allows us to systematically examine how well visual outputs align with the LLM's understanding space. In light of this, we construct a fine-tuning dataset, termed the \textbf{M}ulti-\textbf{G}ranularity \textbf{L}andmark \textbf{D}ataset (MGLD), comprising approximately 200k samples from the GLDv2~\citep{weyand2020GLDv2} dataset, following the process illustrated in ~\Cref{fig:pipeline} (detailed in  
~\Cref{appendix: GLDv2 Dataset}).
In this section, we focus solely on image-text pairs \( (I_i, T_i) \)  in Stage 1 and Q-A pairs \((q_i, a_i)\) in Stage 2.

Specifically, for Stage 1, each landmark in the original dataset corresponds to multiple images, and we pair the landmark name \(T_i\) with the image \(I_i\) that has the highest CLIP similarity to its corresponding landmark names. This selection process ensures that each chosen image aligns closely with the landmark identity.
In Stage 2, we generate Question-Answer (Q-A) pairs to support landmark recognition. We randomly select a question from a predefined set and provide it to GPT-4o along with the selected image and corresponding landmark name. The model then responds with answers describing the landmark. 
Then, the dataset used in this section can be denoted as $D = \{(I_i, T_i, q_i, a_i)\}_{i=1}^{N}$, where \(I_i\) represents the image, \(q_i\) is a question related to the landmark, and \(a_i\) is the corresponding answer, including both image descriptions and location-related information. \(T_i\) is the ground-truth landmark name(\eg, ``Eiffel Tower”).

\paragraph{Measuring knowledge of vision encoder.}
We calculate the CLIP similarity between the visual representation of each image \(I_i\) and all the landmark names \(T_j\) (\eg, ``Eiffel Tower'') in the dataset, using the CLIP vision and text encoders. The similarity, denoted as \(\text{Sim}_{\text{CLIP}}(I_i, T_j)\), is computed as follows:
\vspace{-0.5em}
\begin{equation}
    \text{Sim}_{\text{CLIP}}(I_i, T_j) = \frac{\langle f_v(I_i), f_t(T_j) \rangle}{\|f_v(I_i)\| \|f_t(T_j)\|}
    \label{eq:sim_metric}
\end{equation}
where \(f_v(I_i)\) and \(f_t(T_j)\) represent the visual and textual embeddings of the image and the landmark name, respectively.

We record the {\bf Similarity Score} between the image \(I_i\) and its corresponding ground-truth label \(T_i\), denoted as \(\text{Sim}_{\text{CLIP}}^i\). We also determine the {\bf Relative Similarity Rank (RSR)} of the ground-truth \(T_i\) among all the landmark candidates \(T_j\) for the image \(I_i\), based on the similarity scores, denoted as \(\text{RSR}^i\), indicating the discriminative power of the visual representation. 

We apply various data selection methods to the full dataset based on \(\text{Sim}_{\text{CLIP}}^i\) and \(\text{RSR}^i\), each aimed at capturing specific characteristics that indicate whether the Vision Encoder is in a Known or Unknown state. For consistency, all subsets are of equal size.
\begin{itemize}
   \item \textbf{High Discrimination Selection (HDS):} This method selects images with very high \(\text{RSR}^i\) values, capturing instances where the model effectively distinguishes the ground-truth \(T_i\) from other candidates. HDS emphasizes strong visual discrimination in the VE’s representations, aligning with ``\textit{VE-Known}" characteristics.

   \item \textbf{High Similarity Selection (HSS):} This method focuses on images with high \(\text{Sim}_{\text{CLIP}}^i\) values, highlighting the VE's capacity for capturing rich visual features. HSS prioritizes visual richness over discriminative power, aligning with general ``\textit{VE-Known}" characteristics.

   \item \textbf{Low Clarity Selection (LCS):} This method selects images with both low \(\text{Sim}_{\text{CLIP}}^i\) and low \(\text{RSR}^i\) values, targeting visually ambiguous cases where the model struggles with feature extraction and differentiation. These instances reflect ``\textit{VE-Unknown}" characteristics, indicating limited VE understanding.

   \item \textbf{Balanced Reference Selection (BRS):} This method randomly selects images to serve as a comparative baseline.
\end{itemize}

\paragraph{Evaluation metric.} 
To evaluate LVLM performance, we generate five inference outputs for each question in the test dataset. These responses are assessed across four recognition levels based on their accuracy and detail relative to the correct answer by GPT-4o, with specific examples provided in ~\Cref{appendix: evaluation and statistics}.

\begin{itemize}
    \item \textbf{Strongly Known.} The model consistently delivers detailed, accurate information that closely aligns with the ground truth across multiple responses.
    
    \item \textbf{Known.} At least one response is correct and includes reasonable explanations.
    
    \item \textbf{Weakly Unknown.} The correct entity is not identified, but responses provide relevant hints, such as geographic, architectural, or contextual cues, suggesting an association with the target landmark. 
    
    \item \textbf{Unknown.} None of the responses accurately identify the target. Information is either unrelated or overly generic, lacking specific clues about the target’s identity.
\end{itemize}

The evaluation metrics are defined as follows: each recognition level is quantified by its proportion within the test dataset. \textbf{Accuracy} is defined as the proportion of recognized responses (\textit{i.e.,} the sum of Strongly Known and Known) in the test dataset, representing the model's overall recognition capability.

\begin{figure}[t!]
    \centering
    \includegraphics[width=1\linewidth]{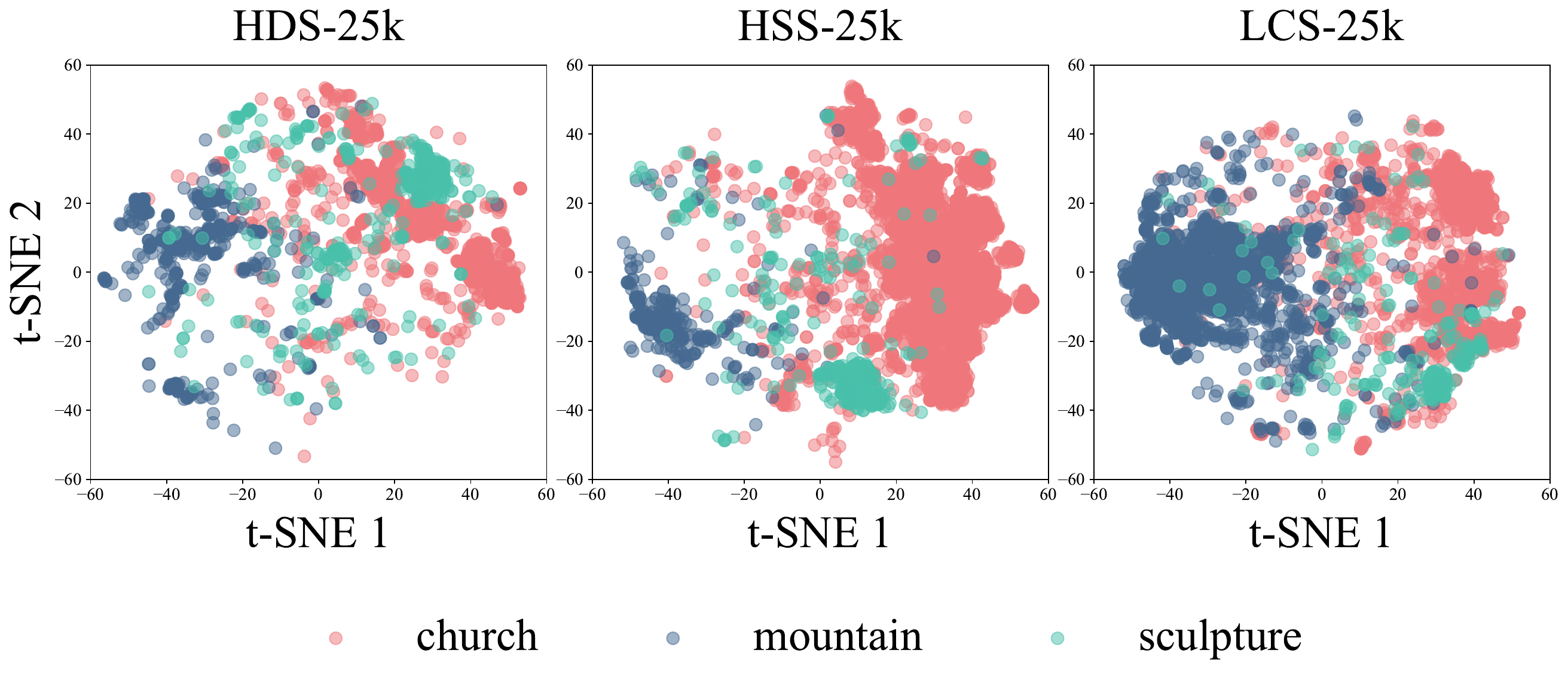}
    \caption{t-SNE visualization of image features. \textbf{Left}: The HDS subset shows more dispersed representations for categories(\eg, ``church''). \textbf{Middle}: The HSS subset shows distinct inter-class separations. \textbf{Right}: The LCS subset shows reduced intra-class variability and less distinct inter-class separations.}
    \label{fig:t-sne}
\end{figure}

\begin{figure}[t!]
    \centering
    \includegraphics[width=\linewidth]{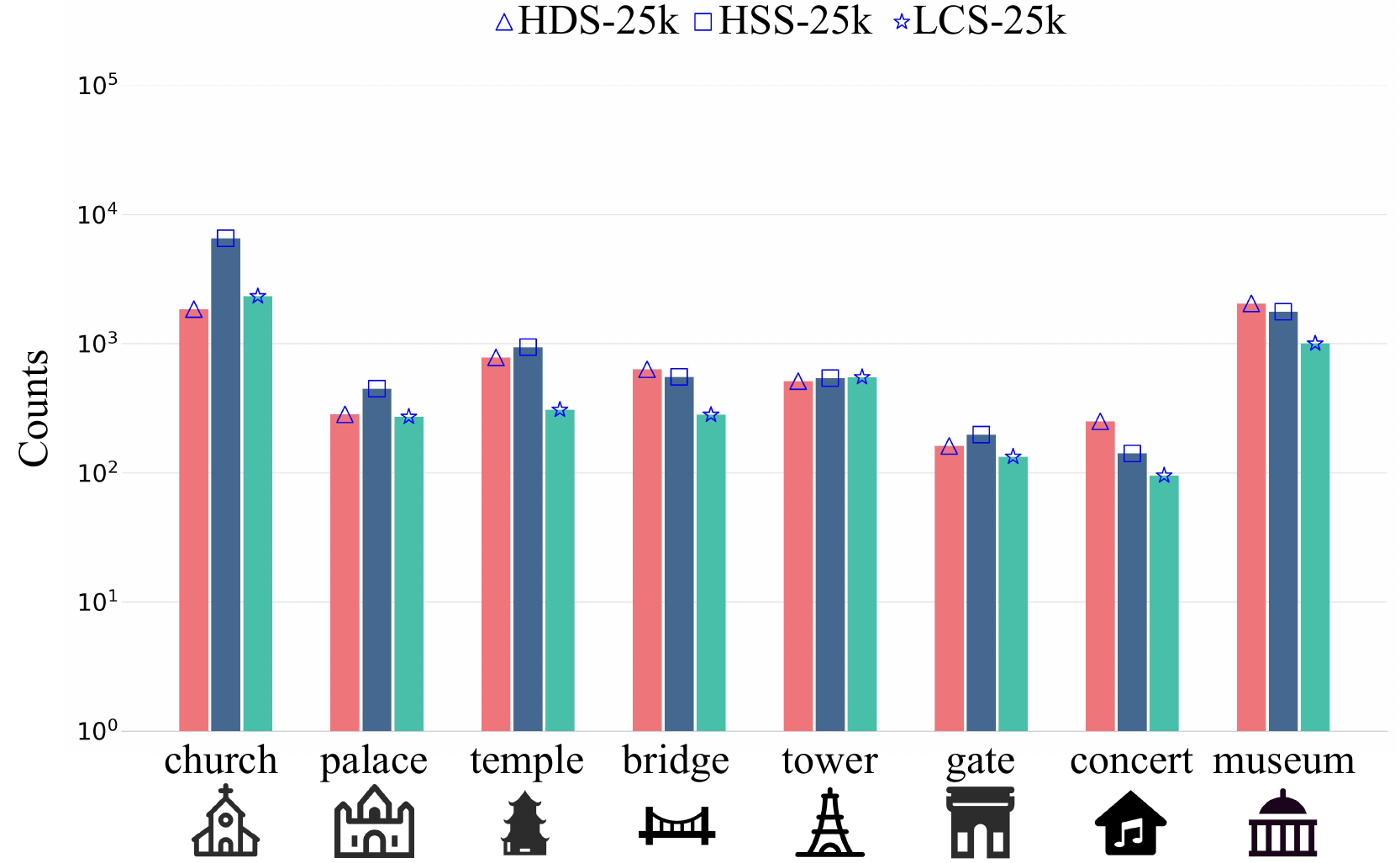}
    \caption{Category counts across subsets. }
    \label{fig:category_log_counts}
    \vspace{-0.5em}
\end{figure}
\subsection{Visual patterns of different knowledge}
Firstly, we analyze the characteristics of different types of visual knowledge, focusing on how variations in visual representation affect cognitive alignment between VE and LLM. As shown in~\Cref{fig:t-sne}, we used t-SNE to visualize the visual knowledge patterns extracted by the VE across distinct subsets. Specifically, we constructed three datasets: HDS-25k (top 25k samples with the highest \(\text{RSR}^i\)), HSS-25k (top 25k samples with the highest \(\text{Sim}_{\text{CLIP}}^i\)), and LCS-25k (25k samples with both low \(\text{RSR}^i\) and low \(\text{Sim}_{\text{CLIP}}^i\)). Additionally,~\Cref{fig:category_log_counts} shows the distribution of major categories in each subset, offering insights into their visual class composition.

The t-SNE results reveal distinct patterns across these subsets. The HDS subset shows dispersed visual representations within categories, indicating that the visual encoder captures fine-grained intra-class diversity, which enhances subtle landmark recognition. Conversely, the HSS subset forms compact, well-separated clusters, suggesting a focus on broad, shared patterns that aid in distinguishing distinct categories like ``mountain" and ``sculpture". The HSS subset also shows the most variation in category counts (see~\Cref{fig:category_log_counts}), supporting its strong inter-class separability. The LCS subset displays reduced intra-class variability and less distinct inter-class boundaries, suggesting challenges in category differentiation.
\definecolor{darkgreen}{rgb}{0.0, 0.5, 0.0}
\definecolor{lightgray}{gray}{0.9}
\definecolor{MidGreen}{HTML}{AAFFAA}
\definecolor{Green}{HTML}{77DD77}
\newcommand{\plusvalue}[1]{\hspace{0.3em}\textcolor{darkgreen}{(+#1)}}
\newcommand{\minusvalue}[1]{\hspace{0.3em}\textcolor{red}{(-#1)}}

\begin{table}[htbp]
    \centering
    \small
    \setlength\tabcolsep{3pt} 

    \begin{tabular}{llll}
        \toprule
       \multirow{2}{*}{\textbf{Category}} &  \multirow{2}{*}{\textbf{Selection Method}} & \multicolumn{2}{l}{\textbf{Data Size}} \\
        \cmidrule(lr){3-4}
        && {25k} & {50k} \\
        \midrule
        \cellcolor{lightgray}{Reference}&\cellcolor{lightgray}BRS & \cellcolor{lightgray}40.1  & \cellcolor{lightgray}56.2  \\
        \midrule
        \textbf{VE-Unknown} & {LCS} & 23.0 {\scriptsize\textbf{\minusvalue{-17.1}}} & 28.1 {\scriptsize\textbf{\minusvalue{-28.1}}} \\
        \midrule
        \textbf{VE-Known}&{HSS} & 40.1  & 60.4 {\scriptsize\textbf{\plusvalue{4.2}}} \\
        \textbf{VE-Known}&HDS & \textbf{49.3} {\scriptsize\textbf{\plusvalue{9.2}}} & \textbf{64.1} {\scriptsize\textbf{\plusvalue{7.9}}} \\
        \bottomrule
    \end{tabular}
    \caption{The baseline accuracy for non-extra data is 8.68\%. The values in the table represent the percentage increase in test accuracy over this baseline, with values in parentheses indicating changes relative to BRS. VE-Unknown shows reduced performance compared to BRS, while VE-Known methods improve accuracy, with HDS achieving the highest increase.}
    \label{tab:data_selection_performance}
\end{table}

\subsection{Uncovering the impact of visual knowledge}
After analyzing the visual patterns of different knowledge extracted by the VE, we investigate a key question: \textit{how do the strength and quality of VE representations influence cognitive alignment with the LLM’s understanding?} To answer this, we evaluated models trained with various data selection strategies, starting from the LLaVA~\citep{liu2023improved} initialization and optimizing both the LLM and the adapter. Strong alignment—indicated by a higher proportion of recognized responses (sum of Strongly Known and Known) — suggests that visual features are both robust and supportive of cross-modal cognitive integration.

\begin{figure}
    \centering
    \includegraphics[width=\linewidth]{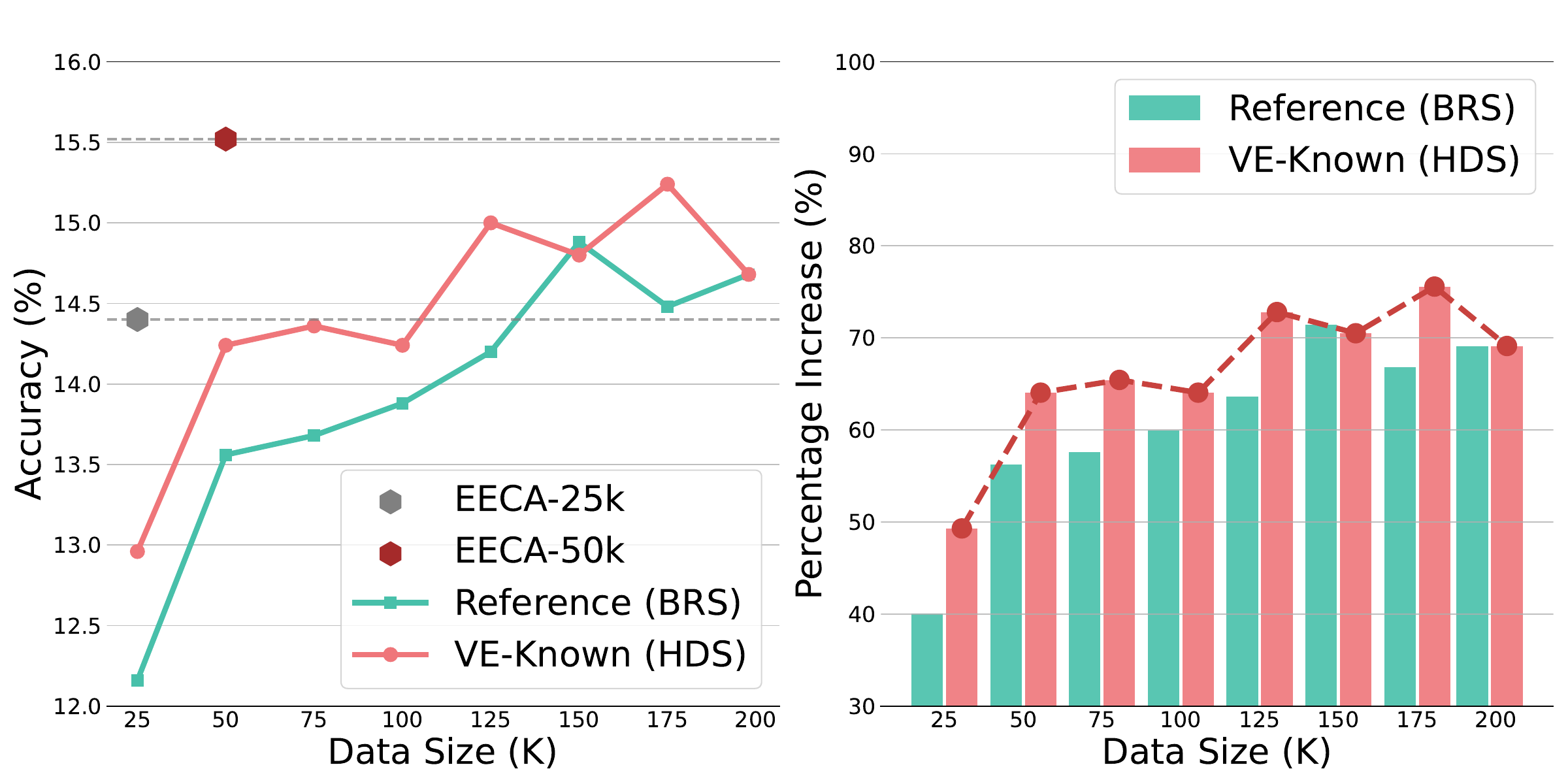}
    \caption{Comparative performance of HDS and BRS selection methods across different dataset sizes. \textbf{Left}: Accuracy (\%) vs. Data Size for BRS and HDS with two point of EECA at 25k and 50k(best). \textbf{Right}: Percentage increase over baseline. VE-Known data outperforms Reference, demonstrating its effectiveness.}
    \label{fig:data_increase_line}
\end{figure}

To measure improvement, we compared the proportion of recognized responses to the LLaVA baseline, focusing on the percentage increase in accurate landmark identifications. Our findings, presented in~\Cref{tab:data_selection_performance} and~\Cref{fig:data_increase_line}, reveal three insights:

\begin{enumerate}
    \item \textbf{VE-Known data mitigates cognitive misalignment.}  
    As shown in~\Cref{tab:data_selection_performance}, VE-Known subsets (HDS and HSS) consistently outperform BRS, especially with smaller datasets. The t-SNE analysis indicates that these subsets have richer, more discriminative visual features, facilitating smoother knowledge transfer from the VE to the LLM.
    
    \item \textbf{VE-Unknown data exacerbates cognitive misalignment.}  
    The Low Clarity Set (LCS), which includes VE-Unknown instances with ambiguous representations, shows the lowest performance (see
    ~\Cref{tab:data_selection_performance}). While VE-Unknown data may contribute partially to LVLM learning, its low confidence and information loss hinder effective cognitive integration, underscoring the need to address VE-Unknown data for better model performance.
    
    \item \textbf{Diminishing returns from increased data volume with VE-Unknown samples.}  
    As shown in~\Cref{fig:data_increase_line}, performance gains plateau with larger datasets, especially when VE-Unknown samples from LCS are included, indicating that data quality has a greater impact than data volume beyond a certain threshold.
\end{enumerate}

Our experiment demonstrates that models trained on quality-driven selections (\eg, HDS), can achieve competitive or superior performance with reduced data sizes. This finding highlights a crucial insight from the perspective of the VE’s cognitive framework: to strengthen cognitive alignment between the VE and the LLM, the information passed from the adapter must be both discriminative and richly informative, helping the LLM distinguish landmarks with greater accuracy. Building on this insight, the following section introduce a method that leverages entity-aware supervision to mitigate cognitive integration challenges in LVLMs.

\section{Open the Eyes of LVLM}\label{sec:method}
\begin{figure*}[htbp]
    \centering
    \includegraphics[width=\linewidth]{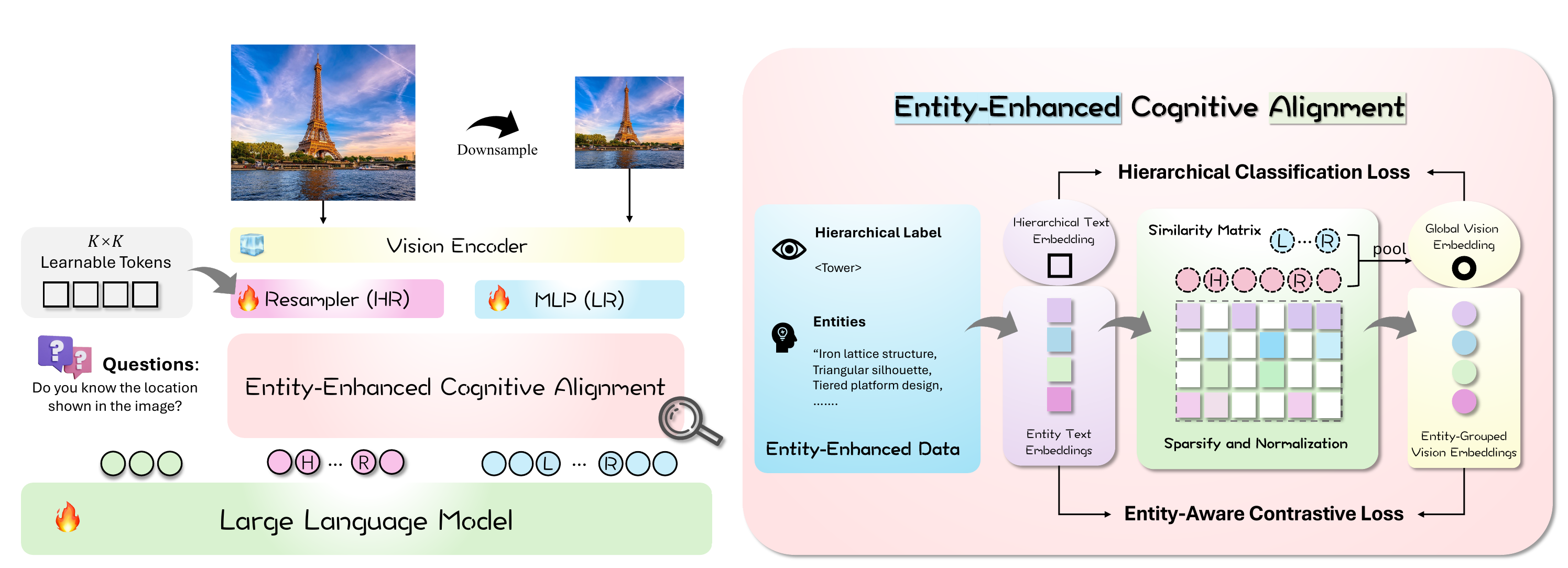}
    \caption{Overview of our model's framework. \ours{} approach combines high- and low-resolution visual features through a dual-branch architecture, supporting alignment with the language model’s cognitive framework. The framework includes a hierarchical classification loss and an entity-aware contrastive loss to encourage rich, discriminative representations aligned with entity-specific information.}
    \label{fig:method}
    \vspace{-0.5em}
\end{figure*}
After examining the cognitive framework of the VE, a natural question arises: \textit{how can we fully ``open the eyes” of the LVLM to achieve cognitive alignment between visual and language components?} In this section, we take initial steps toward answering this question. First, we design a data annotation pipeline to ensure consistency between the visual input and the language model’s output (see~\Cref{subsec:data}). Building on this, we introduce the Entity-Enhanced Cognitive Alignment (\ours{}) framework, which supervises the adapter’s visual tokens to retain richness and discriminative power, minimizing information loss (see~\Cref{subsec:eeca}). This approach encourages transformed visual tokens to ``mimic" VE-Known representations, facilitating alignment with the LLM’s cognitive framework.

\subsection{Multi-granularity data annotation pipeline}\label{subsec:data}
To construct a landmark instruction dataset that enhances the model’s recognition and differentiation capabilities, we designed a 3-stage data pipeline as shown in~\Cref{fig:pipeline}. Stage 1 and 2 have been introduced in ~\Cref{subsec:setup}. Here, we focus on providing a detailed description of Stage 3. 

The final stage, Multi-granularity data annotation, enriches each landmark with hierarchical labels and unique entities, providing a layered structure that captures both broad classifications and fine-grained details. 
Firstly, each landmark is assigned a hierarchical label (\eg, ``mountain," ``lake," ``church,"), placing it within a general category. 
In contrast, entities provides a finer level of detail, capturing landmark-specific attributes. For instance, the Eiffel Tower might be associated with entities like ``iron lattice structure'' and ``triangular silhouette'', which highlight its distinctive physical attributes. These entities are annotated based on both image content and QA pairs, making them directly relevant to the LLM’s answer outputs. 

The resulting dataset, which we denote as $D = \{ (I_j, q_j, a_j, h_j, e_j) \}_{j=1}^{N}$, is structured such that each entry contains an image \(I_j\), a question \(q_j\), an answer \(a_j\) incorporating both descriptive and location-specific information, a hierarchical label \(h_j\) representing the broader category, and unique entities \(e_j\) detailing fine-grained, landmark-specific attributes. 
This dataset construction enables \ours{} to capture the unique identity of each landmark through visual tokens, facilitating cognitive alignment between the vision encoder and the language model.

\subsection{Entity-Enhanced cognitive alignment}\label{subsec:eeca}
\paragraph{Architecture overview.} Within this enriched dataset, we train our model using \ours{}, as illustrated in~\Cref{fig:method}. \ours{} promotes cognitive alignment by supervising the transformation from visual patches to tokens, ensuring that the tokens retain rich, discriminative information linked to specific entities in MGLD. These entities are annotated to align with the LLM’s cognitive framework, supporting cross-modal understanding.

Inspired by recent work~\cite{li2024mini,shi2025we}, we employ a dual-branch visual architecture to leverage high-resolution (HR) information, enhancing the richness of visual representations. Starting with a high-resolution image \( I^H \), we generate a low-resolution version \( I^L \) via bilinear interpolation. \( I^H \) is divided into four sub-images, which, along with \( I^L \), are fed to a shared vision encoder, producing low-resolution and high-resolution visual features, \( V^L \) and \( V^H \).

In the low-resolution branch, a 2-layer MLP adapter outputs visual tokens \( X_v^L \in \mathbb{R}^{N_{v_L} \times C} \). In the high-resolution branch, a shared perceiver resampler~\citep{flamingo} generates high-resolution visual tokens \( X_v^H \in \mathbb{R}^{N_{v_H} \times C} \). To make HR information efficient, \ours{} supervises the compression of HR features, aligning visual tokens with the LLM's embedding space. The LLM then integrates textual and visual information for comprehensive understanding. The EECA framework is optimized with two loss functions: Entity-Aware Contrastive Loss and Hierarchical Classification Loss, detailed below.

\paragraph{Entity-Aware contrastive loss.} 
Given the concatenated visual tokens \( X_v = [X_v^L; X_v^H] \), where \( X_v^H = (X_{v_1}^H, \dots, X_{v_{N_{v_H}}}^H) \in \mathbb{R}^{N_{v_H} \times C} \) represents high-resolution tokens and \( X_v^L \) represents low-resolution tokens, this loss function encourages the model to incorporate fine-grained, entity-specific details from \( X_v^H \) into the primary representation in \( X_v^L \).

Motivated by the idea that different queries can capture diverse aspects of visual information~\citep{kar2024brave}, we aim to learn a combination of query embeddings that corresponds to each entity's token. Specifically,  we construct an entity-grouped vision embedding for each entity token \( \phi(e_j) = X_{e_j} \in \mathbb{R}^C \) as a weighted combination of high-resolution tokens \( X_v^H \), based on a sparsified and normalized similarity matrix, inspired by~\citep{bica2024improving}. This approach preserves entity-relevant characteristics over general visual features.

Following~\citep{yin2024sea}, we propose an Entity-Aware Contrastive Loss to align high-resolution tokens with entity-specific information in the LLM’s embedding space. Formally, let \( W_{i,j} \) be the weight matrix based on the similarity between high-resolution tokens \( X_{i,v}^H \) and entity embeddings \( e_{i,j} \), where \( i \) is the \( i \)-th image and \( j \) is the \( j \)-th entity within that image. The entity-grouped visual embedding \( \tilde{X}_{e_{i,j}} \) is defined as:
\vspace{-0.3em}
\begin{equation}
    \tilde{X}_{e_{i,j}} = \sum_{v=1}^{N_{v_H}} W_{i,j} X_{i,v}^H
    \label{eq:similarity}
    \vspace{-0.3em}
\end{equation}

The Entity-Aware Contrastive Loss \( \mathcal{L}_{ec} \) is then calculated as:
\begin{equation}
    \begin{split}
    \mathcal{L}_{e} & = - \frac{1}{2B} \sum_{i=1}^{B} \sum_{j=1}^{E_i} \left( \log \frac{\exp(S(X_{e_{i,j}} , \tilde{X}_{e_{i,j}} )/ \tau)}{\sum_{k=1}^{E_i} \exp(S(X_{e_{i,j}},\tilde{X}_{e_{i,k}})/ \tau)} \right. \\
    & \left. + \log \frac{\exp(S(\tilde{X}_{e_{i,j}}, X_{e_{i,j}})/ \tau)}{\sum_{k=1}^{E_i} \exp(S(\tilde{X}_{e_{i,j}} , X_{e_{i,k}})/ \tau)} \right)
    \end{split}
    \vspace{-0.5em}
    \label{eq:contrastive loss}
\end{equation}
where \( B \) is the batch size, \( E_i \) is the number of entities for the \( i \)-th image, \( S(a,b) = \frac{a \cdot b}{\|a\| \|b\|} \), and \( \tau \) is the temperature.

\paragraph{Hierarchical classification loss.} 
To enhance the model's understanding across multiple levels of granularity, we introduce a hierarchical classification task. Using the hierarchical labels in MGLD (\eg, ``church,” ``lake,”), this task builds between-category distinctions on top of the within-category distinctions captured through entities.

For each image, classification features are generated by concatenating high-resolution and low-resolution visual tokens, \( X_{i,v}^H \) and \( X_{i,v}^L \), followed by average pooling to obtain a comprehensive representation \( \mathbf{h}_i \). This representation is then used to compute the hierarchical classification loss, defined as:
\vspace{-0.5em}
\begin{equation}
    \mathcal{L}_{h} = - \frac{1}{B} \sum_{i=1}^{B} \log P(y_i | \mathbf{h}_i)
    \vspace{-0.5em}
    \label{eq:classification loss}
\end{equation}
where \( P(y_i | \mathbf{h}_i) \) is the predicted probability of the correct class \( y_i \) for the \( i \)-th image.

\paragraph{Overall objective.} 
The total loss function used to train the model combines three components: a standard language modeling loss (\( \mathcal{L}_{g} \)), a hierarchical classification loss (\( \mathcal{L}_h \)), and an Entity-Aware Contrastive Loss (\( \mathcal{L}_e \)), which together support \ours{} learning.

The language modeling loss \( \mathcal{L}_{g} \) encourages accurate text generation based on the visual input and is defined as:
\vspace{-0.5em}
\begin{equation}
    \mathcal{L}_{g} = - \frac{1}{B} \sum_{i=1}^{B} \sum_{t=1}^{T_i} \log P(w_{i,t} | w_{i,<t}, X_{i})
    \vspace{-0.5em}
    \label{eq:generate loss}
\end{equation}
Overall combined objective can be expressed as $\mathcal{L} = \lambda \mathcal{L}_{g} + \mu_e \mathcal{L}_{e} + \mu_h \mathcal{L}_{h}$
where \( \lambda \), \( \mu_e \), and \( \mu_h \) are balancing coefficients for each loss component.
 
This layered learning process supports cognitive alignment between modalities, as enriched and distinct visual tokens better align with the LLM’s cognitive framework. Optimizing this combined loss enables multi-granularity learning.

\section{Experiments}
\label{sec:exp}
In this section, we analyze the effectiveness of \ours{} on landmark recognition tasks. To evaluate the impact of our approach, we conduct experiments on different variations and data subsets, comparing performance across multiple configurations.
\subsection{Experimental setup}
Our experiments are conducted within the LLaVA-1.5~\citep{liu2023improved}, using a dual-branch visual architecture for enhanced feature representation. We jointly optimize both the LLM and the two adapters within the EECA framework. Further experimental details, including hyperparameter configurations, are provided in ~\Cref{appendix: experiment details}.
For data preparation, we construct the dataset according to the process described in~\Cref{fig:pipeline}, creating subsets with varying data selection methods (See~\Cref{subsec:setup}).
\subsection{Results and analysis}
\paragraph{Effectiveness of \ours{}.} We evaluate EECA against three configurations. \textit{Baseline} serves as a reference without entity prompts. \textit{Entity Prompt (Inference Only)} includes entities only during inference, confirming that entity prompts do not negatively impact performance. \textit{Entity Prompt (Training + Inference)} incorporates entities during both training and inference, yielding significantly higher accuracy and indicating alignment with the LLM's cognitive framework. As shown in~\Cref{fig:data_increase_line}, with only 25k data, EECA reaches the performance of the 125k reference dataset, and with 50k data, it achieves the best results. EECA thus provides balanced improvements without extensive reliance on entity prompts during inference, demonstrating effectiveness in enhancing performance without overfitting. Overall, EECA offers a more efficient, focused approach, achieving strong alignment with the LLM’s cognitive framework.

\begin{table}[htbp]
    \centering
    \small
    \setlength\tabcolsep{2pt} 
    \begin{tabular}{lcc}
        \toprule
        \textbf{Method} & \textbf{Strongly Known} & \textbf{Known} \\
        \midrule
        Baseline & 4.12 & 4.56 \\
        Entity Prompt (Inference Only) & 4.56 & 3.24 \\
        Entity Prompt (Training + Inference) & 19.52 & 9.32 \\
        EECA & 8.52 & 7.0 \\
        \bottomrule
    \end{tabular}
    \caption{Comparison of different methods.}
    \label{tab:main_results}
\end{table}

\paragraph{Ablation study.}
\begin{table}[t]
\centering
\setlength{\tabcolsep}{5pt}
\renewcommand{\arraystretch}{1.0}
\scalebox{0.85}{
\begin{tabular}{lccc}
\toprule
\textbf{Method} & \textbf{Strongly Known} & \textbf{Known} & \textbf{Accuracy} \\
\midrule
Baseline & 4.12 & 4.56 & 8.68 \\
\midrule
\cellcolor{mygreen}
\textit{+ HSS-50k} & 7.48 \plusvalue{3.36} & 6.44 \plusvalue{1.88} & 13.92 \plusvalue{5.24} \\
\midrule
\cellcolor{mygreen}
\textit{+ HR Branch} & 7.92 \plusvalue{0.44} & 5.96 \minusvalue{0.48} & 13.88 \minusvalue{0.04} \\
\midrule
\cellcolor{mygreen}
\textit{+ $\mathcal{L}_{e}$} & 8.48 \plusvalue{0.56} & 5.92 \minusvalue{0.04} & 14.4 \plusvalue{0.52} \\
\midrule
\cellcolor{mygreen}
\textit{+ $\mathcal{L}_{h}$} & \textbf{8.52} \plusvalue{0.04} & \textbf{7.00} \plusvalue{1.08} & \textbf{15.52} \plusvalue{1.12} \\
\bottomrule
\end{tabular}}
\caption{Ablation Study Results.}
\label{tab:ablation}
\end{table}
~\Cref{tab:ablation} demonstrates that the main accuracy gains come from the proposed entity-aware contrastive loss ($\mathcal{L}_e$) and hierarchical classification loss ($\mathcal{L}_h$). In contrast, adding the high-resolution (HR) branch alone does not improve performance without targeted supervision. These results highlight that EECA’s supervision allows the HR branch to deliver richer, more discriminative visual information, aligning the visual encoder’s outputs with the language model’s cognitive framework and enhancing interpretability in landmark recognition.

\subsection{Generalizability of EECA}
To validate EECA’s generalizability, we applied it to three distinct data subsets representing different levels of visual knowledge. Results in~\Cref{tab:eeca_generalizability} show that EECA consistently improves performance across all subsets, demonstrating adaptability even in challenging VE-Unknown scenarios.

For LCS-25k (VE-Unknown data), adding the HR branch significantly boosts performance, indicating that enriched visual information in the HR branch effectively reduces visual ambiguity by enhancing feature richness and separability. In contrast, HDS-25k and HSS-25k (VE-Known data) achieve the largest gains with the entity-aware contrastive loss ($\mathcal{L}_e$) and hierarchical classification loss ($\mathcal{L}_h$), showing that VE-Known data benefit from targeted supervision to extract discriminative features.

Overall, these results underscore EECA’s robustness and adaptability, addressing visual ambiguity in VE-Unknown while enhancing feature discrimination in VE-Known.
\begin{table}[t]
\centering
\setlength{\tabcolsep}{5pt}
\renewcommand{\arraystretch}{1.0}
\scalebox{0.85}{
\begin{tabular}{lccc}
\toprule
\textbf{Method} & \textbf{HDS-25k} & \textbf{HSS-25k} & \textbf{LCS-25k} \\
\midrule
Baseline & 8.68 & 8.68 & 8.68 \\
\midrule
\cellcolor{mygreen}
\textit{+ 25k Data} & 13.00 \plusvalue{4.32} & 13.00 \plusvalue{4.32} & 10.68 \plusvalue{2.00} \\
\midrule
\cellcolor{mygreen}
\textit{+ HR Branch} & 13.60 \plusvalue{4.92} & 13.84 \plusvalue{5.16} & 12.08 \plusvalue{3.40} \\
\midrule
\cellcolor{mygreen}
\textit{+ $\mathcal{L}_{e}$} & 14.00 \plusvalue{5.32} & \textbf{14.40} \plusvalue{5.72} & \textbf{12.32} \plusvalue{3.64} \\
\midrule
\cellcolor{mygreen}
\textit{+ $\mathcal{L}_{h}$} & \textbf{14.40} \plusvalue{5.72} & 13.84 \plusvalue{5.16} & \textbf{12.32} \plusvalue{3.64} \\
\bottomrule
\end{tabular}}
\caption{Performance comparison across VE-Known (HDS, HSS) and VE-Unknown (LCS) subsets.}
\label{tab:eeca_generalizability}
\vspace{-0.5em}
\end{table}

\section{Related Work}
\label{sec:related work}
\paragraph{Large vision language models.}
The integration of vision and language models has advanced Large Vision Language Models for enhanced cross-modal understanding. Foundational models like CLIP~\cite{radford2021learningtransferablevisualmodels} demonstrated the potential of cross-modal applications, while recent models such as Flamingo~\cite{flamingo}, BLIP-2~\cite{blip2}, and LLaVA~\cite{liu2023visual,liu2023improved} refine alignment through efficient strategies and large-scale data. Specialized models like Shikra~\citep{chen2023shikra}, which outputs spatial coordinates in natural language, and Mini-Gemini~\citep{li2024mini}, which employs an additional visual encoder for high-resolution enhancement, further expand functionality. Qwen2-VL~\citep{qwen2.5} introduces a dynamic resolution mechanism for adaptive tokenization across varying image resolutions. Despite these advancements, achieving cognitive alignment between VE and LLM remains challenging.

\paragraph{Visual challenges in LVLMs.} Despite significant progress, many visual challenges remain due to inherent limitations in vision encoders like CLIP. While widely adopted, CLIP often fails to capture comprehensive visual information, hindering LVLM performance. Research~\citep{wei2025vary} shows that CLIP struggles with various visual tasks, reducing the effectiveness of CLIP-based LVLMs. Another study~\citep{tang2023lemons} suggests that CLIP may treat visual inputs as a “bag of concepts,” missing higher-level structures like part-whole relationships. Furthermore,~\citep{tong2024eyes} found that LVLMs often fail on simple questions due to CLIP’s pre-training limitations, which overlook critical visual details and struggle to prioritize significant patterns. Additionally, the vision encoder's performance is constrained by resolution and data quality~\cite{zhang2024internlm,qwen2.5}. These studies underscore the need for techniques to fully unlock the potential of LVLMs.

\section{Conclusion and Discussion}
\label{sec:conlusion}
This study revisits the question: does seeing always mean knowing? In Large Vision Language Models (LVLMs), we find that this is often not the case. Our investigation reveals a critical cognitive misalignment between the vision encoder (VE) and the large language model (LLM), where VE’s visual representations do not fully align with the LLM’s cognitive framework. Our results show that LVLM performance is closely tied to the knowledge within the VE: VE-Known data alleviate cognitive misalignment, while VE-Unknown data exacerbate it. To address this gap, we propose Entity-Enhanced Cognitive Alignment (EECA), which supervises the adapter’s visual tokens to retain richness and discriminative power through multigranular representation, enabling them to “mimic” VE-Known behavior. Our results demonstrate that EECA is effective, robust, and generalizable, improving performance across both VE-Known and VE-Unknown data.

Despite these advancements, EECA has certain limitations. Its effectiveness relies on labeled data to generate supervision signals, limiting its applicability to tasks with available annotations. Additionally, EECA’s design focuses on entity recognition and does not provide a general solution for cognitive misalignment beyond this scope. We hope our findings inspire future work toward broader solutions for cognitive alignment in multimodal systems.

{
    \small
    \bibliographystyle{ieeenat_fullname}
    \bibliography{main}

\begin{thebibliography}{45}
\providecommand{\natexlab}[1]{#1}
\providecommand{\url}[1]{\texttt{#1}}
\expandafter\ifx\csname urlstyle\endcsname\relax
  \providecommand{\doi}[1]{doi: #1}\else
  \providecommand{\doi}{doi: \begingroup \urlstyle{rm}\Url}\fi

\bibitem[AI@Meta(2024)]{llama3modelcard}
AI@Meta.
\newblock Llama 3 model card.
\newblock 2024.

\bibitem[Alayrac et~al.(2022)Alayrac, Donahue, Luc, Miech, Barr, Hasson, Lenc, Mensch, Millican, Reynolds, et~al.]{flamingo}
Jean-Baptiste Alayrac, Jeff Donahue, Pauline Luc, Antoine Miech, Iain Barr, Yana Hasson, Karel Lenc, Arthur Mensch, Katherine Millican, Malcolm Reynolds, et~al.
\newblock Flamingo: a visual language model for few-shot learning.
\newblock In \emph{NeurIPS}, 2022.

\bibitem[Bai et~al.(2023)Bai, Bai, Yang, Wang, Tan, Wang, Lin, Zhou, and Zhou]{bai2023qwen}
Jinze Bai, Shuai Bai, Shusheng Yang, Shijie Wang, Sinan Tan, Peng Wang, Junyang Lin, Chang Zhou, and Jingren Zhou.
\newblock Qwen-vl: A frontier large vision-language model with versatile abilities.
\newblock \emph{arXiv preprint arXiv:2308.12966}, 2023.

\bibitem[Bica et~al.(2024)Bica, Ili{\'c}, Bauer, Erdogan, Bo{\v{s}}njak, Kaplanis, Gritsenko, Minderer, Blundell, Pascanu, et~al.]{bica2024improving}
Ioana Bica, Anastasija Ili{\'c}, Matthias Bauer, Goker Erdogan, Matko Bo{\v{s}}njak, Christos Kaplanis, Alexey~A Gritsenko, Matthias Minderer, Charles Blundell, Razvan Pascanu, et~al.
\newblock Improving fine-grained understanding in image-text pre-training.
\newblock \emph{arXiv preprint arXiv:2401.09865}, 2024.

\bibitem[Chen et~al.(2023)Chen, Zhang, Zeng, Zhang, Zhu, and Zhao]{chen2023shikra}
Keqin Chen, Zhao Zhang, Weili Zeng, Richong Zhang, Feng Zhu, and Rui Zhao.
\newblock Shikra: Unleashing multimodal llm's referential dialogue magic.
\newblock \emph{arXiv preprint arXiv:2306.15195}, 2023.

\bibitem[Dai et~al.(2023)Dai, Li, Li, Tiong, Zhao, Wang, Li, Fung, and Hoi]{dai2023instructblip}
Wenliang Dai, Junnan Li, D Li, AMH Tiong, J Zhao, W Wang, B Li, P Fung, and S Hoi.
\newblock Instructblip: Towards general-purpose vision-language models with instruction tuning. arxiv 2023.
\newblock \emph{arXiv preprint arXiv:2305.06500}, 2, 2023.

\bibitem[Dong et~al.(2024)Dong, Zhang, Zang, Cao, Wang, Ouyang, Zhang, Duan, Zhang, Li, et~al.]{dong2024internlm}
Xiaoyi Dong, Pan Zhang, Yuhang Zang, Yuhang Cao, Bin Wang, Linke Ouyang, Songyang Zhang, Haodong Duan, Wenwei Zhang, Yining Li, et~al.
\newblock Internlm-xcomposer2-4khd: A pioneering large vision-language model handling resolutions from 336 pixels to 4k hd.
\newblock \emph{arXiv preprint arXiv:2404.06512}, 2024.

\bibitem[Gekhman et~al.(2024)Gekhman, Yona, Aharoni, Eyal, Feder, Reichart, and Herzig]{gekhman2024does}
Zorik Gekhman, Gal Yona, Roee Aharoni, Matan Eyal, Amir Feder, Roi Reichart, and Jonathan Herzig.
\newblock Does fine-tuning llms on new knowledge encourage hallucinations?
\newblock \emph{arXiv preprint arXiv:2405.05904}, 2024.

\bibitem[Google(2023)]{Bard}
Google.
\newblock Bard, 2023.

\bibitem[Goyal et~al.(2017)Goyal, Khot, Summers-Stay, Batra, and Parikh]{goyal2017making}
Yash Goyal, Tejas Khot, Douglas Summers-Stay, Dhruv Batra, and Devi Parikh.
\newblock Making the {V} in {VQA} matter: Elevating the role of image understanding in visual question answering.
\newblock In \emph{CVPR}, 2017.

\bibitem[Hudson and Manning(2019)]{hudson2019gqa}
Drew~A Hudson and Christopher~D Manning.
\newblock {GQA}: A new dataset for real-world visual reasoning and compositional question answering.
\newblock In \emph{CVPR}, 2019.

\bibitem[Kar et~al.(2024)Kar, Tonioni, Poklukar, Kulshrestha, Zamir, and Tombari]{kar2024brave}
O{\u{g}}uzhan~Fatih Kar, Alessio Tonioni, Petra Poklukar, Achin Kulshrestha, Amir Zamir, and Federico Tombari.
\newblock Brave: Broadening the visual encoding of vision-language models.
\newblock \emph{arXiv preprint arXiv:2404.07204}, 2024.

\bibitem[Kazemzadeh et~al.(2014)Kazemzadeh, Ordonez, Matten, and Berg]{kazemzadeh2014referitgame}
Sahar Kazemzadeh, Vicente Ordonez, Mark Matten, and Tamara Berg.
\newblock Referitgame: Referring to objects in photographs of natural scenes.
\newblock In \emph{EMNLP}, 2014.

\bibitem[Krishna et~al.(2017)Krishna, Zhu, Groth, Johnson, Hata, Kravitz, Chen, Kalantidis, Li, Shamma, et~al.]{krishna2017visual}
Ranjay Krishna, Yuke Zhu, Oliver Groth, Justin Johnson, Kenji Hata, Joshua Kravitz, Stephanie Chen, Yannis Kalantidis, Li-Jia Li, David~A Shamma, et~al.
\newblock Visual genome: Connecting language and vision using crowdsourced dense image annotations.
\newblock \emph{IJCV}, 2017.

\bibitem[Lauren{\c{c}}on et~al.(2024)Lauren{\c{c}}on, Tronchon, Cord, and Sanh]{laurenccon2024matters}
Hugo Lauren{\c{c}}on, L{\'e}o Tronchon, Matthieu Cord, and Victor Sanh.
\newblock What matters when building vision-language models?
\newblock \emph{arXiv preprint arXiv:2405.02246}, 2024.

\bibitem[Li et~al.(2023)Li, Li, Savarese, and Hoi]{blip2}
Junnan Li, Dongxu Li, Silvio Savarese, and Steven Hoi.
\newblock Blip-2: Bootstrapping language-image pre-training with frozen image encoders and large language models.
\newblock \emph{arXiv:2301.12597}, 2023.

\bibitem[Li et~al.(2024)Li, Zhang, Wang, Zhong, Chen, Chu, Liu, and Jia]{li2024mini}
Yanwei Li, Yuechen Zhang, Chengyao Wang, Zhisheng Zhong, Yixin Chen, Ruihang Chu, Shaoteng Liu, and Jiaya Jia.
\newblock Mini-gemini: Mining the potential of multi-modality vision language models.
\newblock \emph{arXiv preprint arXiv:2403.18814}, 2024.

\bibitem[Liu et~al.(2024)Liu, Zhang, Qiu, Huang, Lin, Zhao, Geng, Lin, Jin, Zhang, et~al.]{liu2024sphinx}
Dongyang Liu, Renrui Zhang, Longtian Qiu, Siyuan Huang, Weifeng Lin, Shitian Zhao, Shijie Geng, Ziyi Lin, Peng Jin, Kaipeng Zhang, et~al.
\newblock Sphinx-x: Scaling data and parameters for a family of multi-modal large language models.
\newblock \emph{arXiv preprint arXiv:2402.05935}, 2024.

\bibitem[Liu et~al.(2023{\natexlab{a}})Liu, Li, Li, and Lee]{liu2023improved}
Haotian Liu, Chunyuan Li, Yuheng Li, and Yong~Jae Lee.
\newblock Improved baselines with visual instruction tuning.
\newblock \emph{arXiv preprint arXiv:2310.03744}, 2023{\natexlab{a}}.

\bibitem[Liu et~al.(2023{\natexlab{b}})Liu, Li, Wu, and Lee]{liu2023visual}
Haotian Liu, Chunyuan Li, Qingyang Wu, and Yong~Jae Lee.
\newblock Visual instruction tuning.
\newblock 2023{\natexlab{b}}.

\bibitem[Loshchilov and Hutter(2017)]{loshchilov2017decoupled}
Ilya Loshchilov and Frank Hutter.
\newblock Decoupled weight decay regularization.
\newblock In \emph{ICLR}, 2017.

\bibitem[Lu et~al.(2024)Lu, Liu, Zhang, Wang, Dong, Liu, Sun, Ren, Li, Yang, et~al.]{lu2024deepseek}
Haoyu Lu, Wen Liu, Bo Zhang, Bingxuan Wang, Kai Dong, Bo Liu, Jingxiang Sun, Tongzheng Ren, Zhuoshu Li, Hao Yang, et~al.
\newblock Deepseek-vl: towards real-world vision-language understanding.
\newblock \emph{arXiv preprint arXiv:2403.05525}, 2024.

\bibitem[Mao et~al.(2016)Mao, Huang, Toshev, Camburu, Yuille, and Murphy]{mao2016generation}
Junhua Mao, Jonathan Huang, Alexander Toshev, Oana Camburu, Alan~L Yuille, and Kevin Murphy.
\newblock Generation and comprehension of unambiguous object descriptions.
\newblock In \emph{CVPR}, 2016.

\bibitem[Marino et~al.(2019)Marino, Rastegari, Farhadi, and Mottaghi]{marino2019ok}
Kenneth Marino, Mohammad Rastegari, Ali Farhadi, and Roozbeh Mottaghi.
\newblock {OK-VQA}: A visual question answering benchmark requiring external knowledge.
\newblock In \emph{CVPR}, 2019.

\bibitem[Mishra et~al.(2019)Mishra, Shekhar, Singh, and Chakraborty]{mishra2019ocr}
Anand Mishra, Shashank Shekhar, Ajeet~Kumar Singh, and Anirban Chakraborty.
\newblock {OCR-VQA}: Visual question answering by reading text in images.
\newblock In \emph{ICDAR}, 2019.

\bibitem[OpenAI(2023)]{openai2023gpt4}
OpenAI.
\newblock Gpt-4 technical report, 2023.

\bibitem[OpenAI(2024)]{gpt4o}
OpenAI.
\newblock {GPT-4o System Card}, 2024.

\bibitem[Peng et~al.(2023)Peng, Li, He, Galley, and Gao]{peng2023instruction}
Baolin Peng, Chunyuan Li, Pengcheng He, Michel Galley, and Jianfeng Gao.
\newblock Instruction tuning with gpt-4.
\newblock \emph{arXiv preprint arXiv:2304.03277}, 2023.

\bibitem[Radford et~al.(2021)Radford, Kim, Hallacy, Ramesh, Goh, Agarwal, Sastry, Askell, Mishkin, Clark, Krueger, and Sutskever]{radford2021learningtransferablevisualmodels}
Alec Radford, Jong~Wook Kim, Chris Hallacy, Aditya Ramesh, Gabriel Goh, Sandhini Agarwal, Girish Sastry, Amanda Askell, Pamela Mishkin, Jack Clark, Gretchen Krueger, and Ilya Sutskever.
\newblock Learning transferable visual models from natural language supervision, 2021.

\bibitem[Schwenk et~al.(2022)Schwenk, Khandelwal, Clark, Marino, and Mottaghi]{schwenk2022okvqa}
Dustin Schwenk, Apoorv Khandelwal, Christopher Clark, Kenneth Marino, and Roozbeh Mottaghi.
\newblock {A-OKVQA}: A benchmark for visual question answering using world knowledge.
\newblock In \emph{ECCV}, 2022.

\bibitem[Sharma et~al.(2018)Sharma, Ding, Goodman, and Soricut]{sharma2018conceptual}
Piyush Sharma, Nan Ding, Sebastian Goodman, and Radu Soricut.
\newblock Conceptual captions: A cleaned, hypernymed, image alt-text dataset for automatic image captioning.
\newblock In \emph{ACL}, 2018.

\bibitem[Shi et~al.(2025)Shi, Wu, Mao, Wang, and Darrell]{shi2025we}
Baifeng Shi, Ziyang Wu, Maolin Mao, Xin Wang, and Trevor Darrell.
\newblock When do we not need larger vision models?
\newblock In \emph{European Conference on Computer Vision}, pages 444--462. Springer, 2025.

\bibitem[Sidorov et~al.(2020)Sidorov, Hu, Rohrbach, and Singh]{sidorov2020textcaps}
Oleksii Sidorov, Ronghang Hu, Marcus Rohrbach, and Amanpreet Singh.
\newblock Textcaps: a dataset for image captioning with reading comprehension.
\newblock In \emph{ECCV}, 2020.

\bibitem[Sun et~al.(2023)Sun, Fang, Wu, Wang, and Cao]{sun2023eva}
Quan Sun, Yuxin Fang, Ledell Wu, Xinlong Wang, and Yue Cao.
\newblock Eva-clip: Improved training techniques for clip at scale.
\newblock \emph{arXiv preprint arXiv:2303.15389}, 2023.

\bibitem[Tang et~al.(2023)Tang, Yamada, Zhang, and Yildirim]{tang2023lemons}
Yingtian Tang, Yutaro Yamada, Yoyo Zhang, and Ilker Yildirim.
\newblock When are lemons purple? the concept association bias of vision-language models.
\newblock In \emph{Proceedings of the 2023 Conference on Empirical Methods in Natural Language Processing}, pages 14333--14348, 2023.

\bibitem[Team(2024)]{qwen2.5}
Qwen Team.
\newblock Qwen2.5: A party of foundation models, 2024.

\bibitem[Tong et~al.(2024)Tong, Liu, Zhai, Ma, LeCun, and Xie]{tong2024eyes}
Shengbang Tong, Zhuang Liu, Yuexiang Zhai, Yi Ma, Yann LeCun, and Saining Xie.
\newblock Eyes wide shut? exploring the visual shortcomings of multimodal llms.
\newblock In \emph{Proceedings of the IEEE/CVF Conference on Computer Vision and Pattern Recognition}, pages 9568--9578, 2024.

\bibitem[Wang et~al.(2024)Wang, Bai, Tan, Wang, Fan, Bai, Chen, Liu, Wang, Ge, et~al.]{wang2024qwen2}
Peng Wang, Shuai Bai, Sinan Tan, Shijie Wang, Zhihao Fan, Jinze Bai, Keqin Chen, Xuejing Liu, Jialin Wang, Wenbin Ge, et~al.
\newblock Qwen2-vl: Enhancing vision-language model's perception of the world at any resolution.
\newblock \emph{arXiv preprint arXiv:2409.12191}, 2024.

\bibitem[Wei et~al.(2025)Wei, Kong, Chen, Zhao, Ge, Yang, Sun, Han, and Zhang]{wei2025vary}
Haoran Wei, Lingyu Kong, Jinyue Chen, Liang Zhao, Zheng Ge, Jinrong Yang, Jianjian Sun, Chunrui Han, and Xiangyu Zhang.
\newblock Vary: Scaling up the vision vocabulary for large vision-language model.
\newblock In \emph{European Conference on Computer Vision}, pages 408--424. Springer, 2025.

\bibitem[Weyand et~al.(2020)Weyand, Araujo, Cao, and Sim]{weyand2020GLDv2}
T. Weyand, A. Araujo, B. Cao, and J. Sim.
\newblock {Google Landmarks Dataset v2 - A Large-Scale Benchmark for Instance-Level Recognition and Retrieval}.
\newblock In \emph{Proc. CVPR}, 2020.

\bibitem[Xue et~al.(2024)Xue, Shu, Awadalla, Wang, Yan, Purushwalkam, Zhou, Prabhu, Dai, Ryoo, et~al.]{xue2024xgen}
Le Xue, Manli Shu, Anas Awadalla, Jun Wang, An Yan, Senthil Purushwalkam, Honglu Zhou, Viraj Prabhu, Yutong Dai, Michael~S Ryoo, et~al.
\newblock xgen-mm (blip-3): A family of open large multimodal models.
\newblock \emph{arXiv preprint arXiv:2408.08872}, 2024.

\bibitem[Ye et~al.(2024)Ye, Xu, Liu, Hu, Yan, Qian, Zhang, Huang, and Zhou]{ye2024mplug}
Jiabo Ye, Haiyang Xu, Haowei Liu, Anwen Hu, Ming Yan, Qi Qian, Ji Zhang, Fei Huang, and Jingren Zhou.
\newblock mplug-owl3: Towards long image-sequence understanding in multi-modal large language models.
\newblock \emph{arXiv preprint arXiv:2408.04840}, 2024.

\bibitem[Yin et~al.(2024)Yin, Zhao, Zhang, Lin, Wang, Tao, Wan, Zhang, Yin, and Zhang]{yin2024sea}
Yuanyang Yin, Yaqi Zhao, Yajie Zhang, Ke Lin, Jiahao Wang, Xin Tao, Pengfei Wan, Di Zhang, Baoqun Yin, and Wentao Zhang.
\newblock Sea: Supervised embedding alignment for token-level visual-textual integration in mllms.
\newblock \emph{arXiv preprint arXiv:2408.11813}, 2024.

\bibitem[Zhai et~al.(2023)Zhai, Mustafa, Kolesnikov, and Beyer]{zhai2023sigmoid}
Xiaohua Zhai, Basil Mustafa, Alexander Kolesnikov, and Lucas Beyer.
\newblock Sigmoid loss for language image pre-training.
\newblock In \emph{Proceedings of the IEEE/CVF International Conference on Computer Vision}, pages 11975--11986, 2023.

\bibitem[Zhang et~al.(2024)Zhang, Dong, Zang, Cao, Qian, Chen, Guo, Duan, Wang, Ouyang, et~al.]{zhang2024internlm}
Pan Zhang, Xiaoyi Dong, Yuhang Zang, Yuhang Cao, Rui Qian, Lin Chen, Qipeng Guo, Haodong Duan, Bin Wang, Linke Ouyang, et~al.
\newblock Internlm-xcomposer-2.5: A versatile large vision language model supporting long-contextual input and output.
\newblock \emph{arXiv preprint arXiv:2407.03320}, 2024.

\end{thebibliography}
}

\newpage
\clearpage
\appendix
\setcounter{page}{1}
\maketitlesupplementary
\section{Experiment Details} \label{appendix: experiment details}
\paragraph{Architecture.} We utilize CLIP-ViT-L-14~\citep{radford2021learningtransferablevisualmodels} as the vision encoder, with a default resolution of 336 × 336, and Meta-Llama-3-8B~\citep{llama3modelcard} as the language model. The low-resolution (LR) branch employs a 2-layer MLP as the adapter, while the high-resolution (HR) branch compresses visual tokens using a shared Perceiver resampler layer~\citep{flamingo}. 
In the HR branch, each high-resolution image is divided into four sub-patches. The resampler processes the visual tokens from these sub-images, compressing them from 2,880 down to 128 tokens via cross-attention with query vectors. These 128 tokens are then concatenated with the 576 tokens from the low-resolution overview image and fed into the LLM. 

\paragraph{Pretrain Datasets.}
We use the same dataset for LLaVA-1.5 experiments. Specifically, stage 1 uses CC595k~\citep{sharma2018conceptual} and stage 2 uses DataMix 665k~\citep{liu2023visual, goyal2017making, hudson2019gqa, marino2019ok, mishra2019ocr, schwenk2022okvqa, sidorov2020textcaps, mao2016generation, kazemzadeh2014referitgame, krishna2017visual} proposed in~\citep{liu2023improved}. 

\paragraph{Hyperparameters.}
In this work, we adopt the same set of hyperparameters as LLaVA-1.5~\citep{liu2023improved}. We show the training hyperparameters for LLaVA-1.5 experiments in Table~\ref{tab:hyperparameters}. All experiments are conducted using a maximum of 8 Nvidia H800 GPUs. We set the value of $\mu_e=7.32$ and $\mu_h=4.38$, respectively, with a sparsification threshold $\theta=0.5$ applied to selectively filter out lower-relevance tokens. Additionally, the temperature parameter, initialized at zero, is set as learnable to dynamically adjust throughout training.  

\begin{figure}[htbp]
    \centering
    \includegraphics[width=\linewidth]{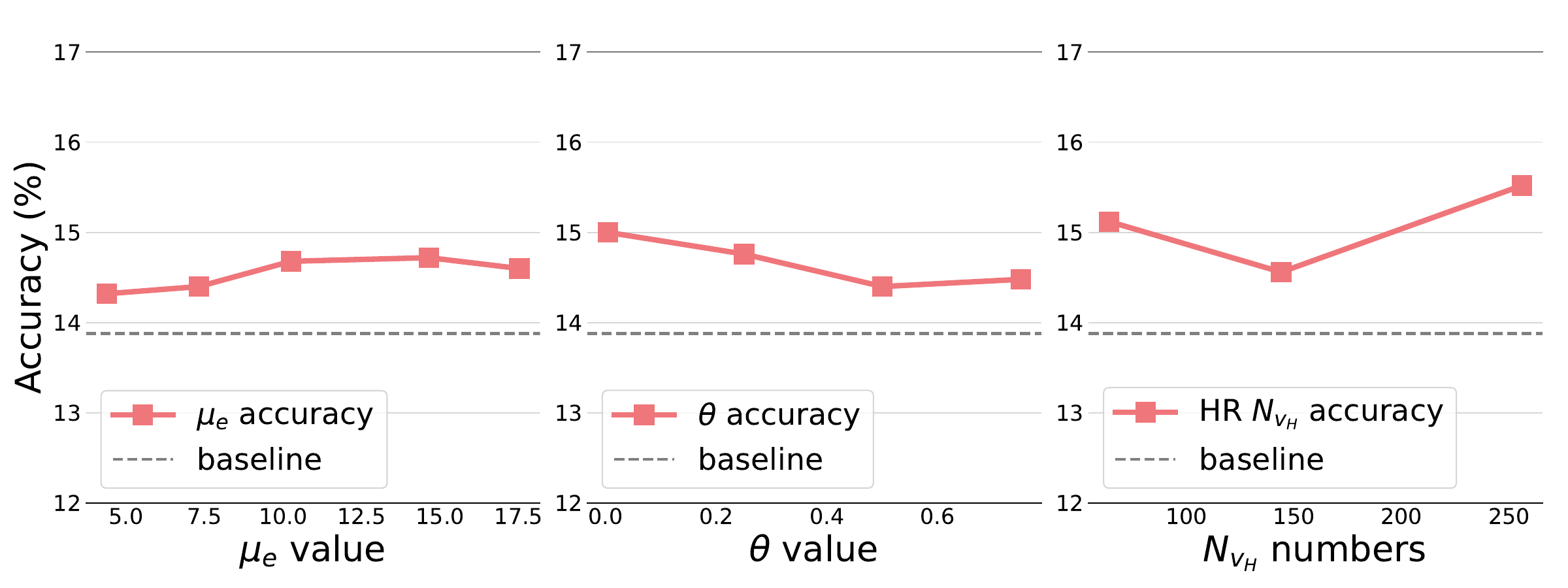}
    \caption{The robustness of the different hyperparameters.}
    \label{fig:parameters}
\end{figure}

\paragraph{Robustness of EECA.}
We evaluate the robustness of EECA across different hyperparameter settings. The model maintains comparable accuracy within specific ranges of key parameters, such as the balancing coefficient $\mu_e$, sparsification threshold $\theta$, and high-resolution visual tokens $N_{v_{H}}$ (see ~\Cref{fig:parameters}), demonstrating stable performance despite small variations.

\begin{table}[htbp]
\centering
\small
\begin{tabular}{@{}lcccc@{}}
\toprule
\multirow{2}{*}{Hyperparameter} & \multicolumn{2}{c}{LLaVA-1.5}  & \multicolumn{1}{c}{EECA} \\
& Stage 1         & Stage 2   & Stage 3   \\ 
\midrule
batch size      & 256             & 128    &128        \\
lr           & 2e-3          & 2e-5    & 2e-5        \\
lr schedule decay   & cosine    & cosine    & cosine               \\
lr warmup ratio      & 0.03   & 0.03        & 0.03          \\
weight decay   & 0        & 0        & 0     \\
epoch   & 1       & 1      & $1^*$      \\
optimizer & \multicolumn{3}{c}{AdamW~\citep{loshchilov2017decoupled}} \\
DeepSpeed stage   & 2      & 2      &2        \\
\bottomrule
\end{tabular}
\caption{Hyperparameters for EECA training on LLaVA-1.5. By default, EECA is trained for 1 epoch, denoted by $*$. Unless otherwise stated, the results presented in ~\Cref{tab:eeca_generalizability} are based on 2 epochs of training.}
\label{tab:hyperparameters}
\end{table}

\section{Datasets Construction for MGLD} \label{appendix: GLDv2 Dataset}
\begin{figure*}[t!]
    \centering
    \includegraphics[width=\linewidth]{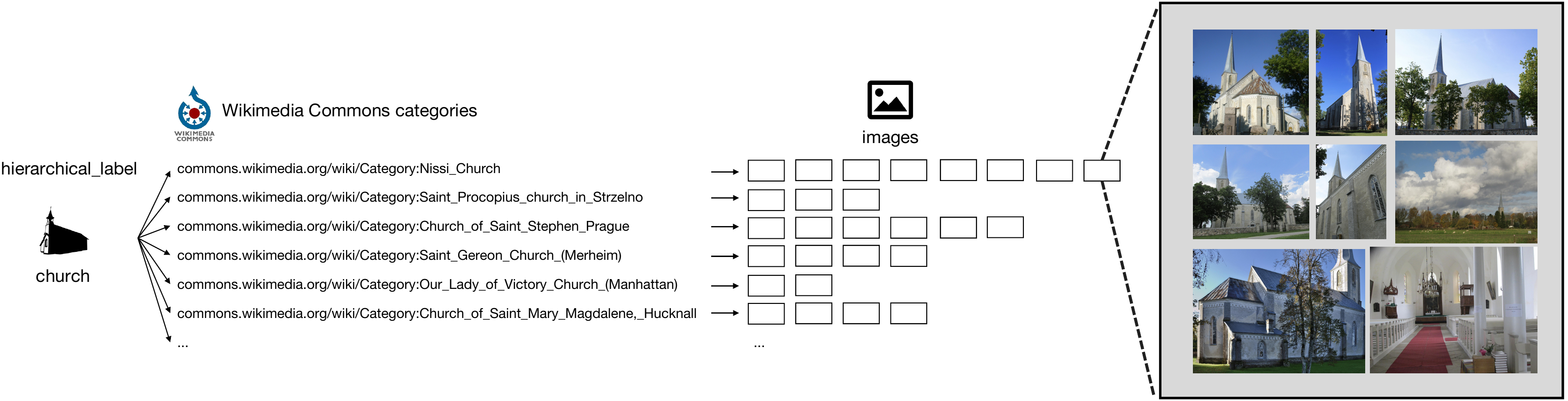}
    \caption{The structure of the GLDv2 train set}
    \label{fig:data_for_structure}
\end{figure*}
This section expands ~\Cref{sec:sight2insight} with additional details about our data preprocessing steps. 
\subsection{Details of google landmarks dataset v2}
\paragraph{Overview.}
The Google Landmarks Dataset v2 (GLDv2)~\citep{weyand2020GLDv2} is the largest benchmark for fine-grained instance recognition and image retrieval, comprising over 5 million images with 200,000 distinct instance labels. It sourced from Wikimedia Commons, and is characterized by real-world challenges such as imbalanced class distribution and high intra-class variability.

\paragraph{Data usage in this work.} In this study, we leverage the GLDv2 dataset to construct a fine-tuning dataset, utilizing all data from the training set (\texttt{train.csv}, \texttt{train\_label\_to\_category.csv}, and \texttt{train\_label\_to\_hierarchical.csv}). This dataset comprises 4.1 million images spanning 203,000 landmarks.

\begin{itemize}
\item 
\textbf{\texttt{train.csv}}: Contains fields \texttt{id}, \texttt{url}, and \texttt{landmark\_id}. Here, \texttt{id} is a 16-character string, \texttt{url} is a string representing the image's URL, and \texttt{landmark\_id} is an integer identifier for the landmark.

\item 
\textbf{\texttt{train\_label\_to\_category.csv}}: Includes \texttt{landmark\_id} and \texttt{category} fields. \texttt{landmark\_id} is an integer, while \texttt{category} is a Wikimedia URL linking to the class definition of the landmark.

\item 
\textbf{\texttt{train\_label\_to\_hierarchical.csv}}: Contains fields \texttt{landmark\_id}, \texttt{category}, \texttt{supercategory}, \texttt{hierarchical\_label}, and \texttt{natural\_or\_human\_made}. \texttt{Supercategory} refers to the type of landmark (e.g., natural or human-made), mined from Wikimedia. \texttt{Hierarchical\_label} corresponds to the landmark's hierarchical classification, and \texttt{natural\_or\_human\_made} indicates whether the landmark is naturally occurring or man-made.
\end{itemize}

The structure of the GLDv2 training dataset is depicted in ~\Cref{fig:data_for_structure}. Each \texttt{hierarchical\_label} encompasses multiple categories, and each category consists of a varying number of images, reflecting the diverse and hierarchical nature of the dataset.
The category distribution in GLDv2 training dataset is highly imbalanced, as illustrated in ~\Cref{fig:distribution_of_category_counts}. Approximately 57\% of the categories contain at most 10 images, and 38\% have 5 or fewer images. This makes the dataset diverse, covering a wide range of landmarks, from globally renowned sites to more obscure, local landmarks.

\subsection{Image selection methodology}\label{subappendix:selection}
\begin{figure}[t!]
    \centering
    \includegraphics[width=0.8\linewidth]{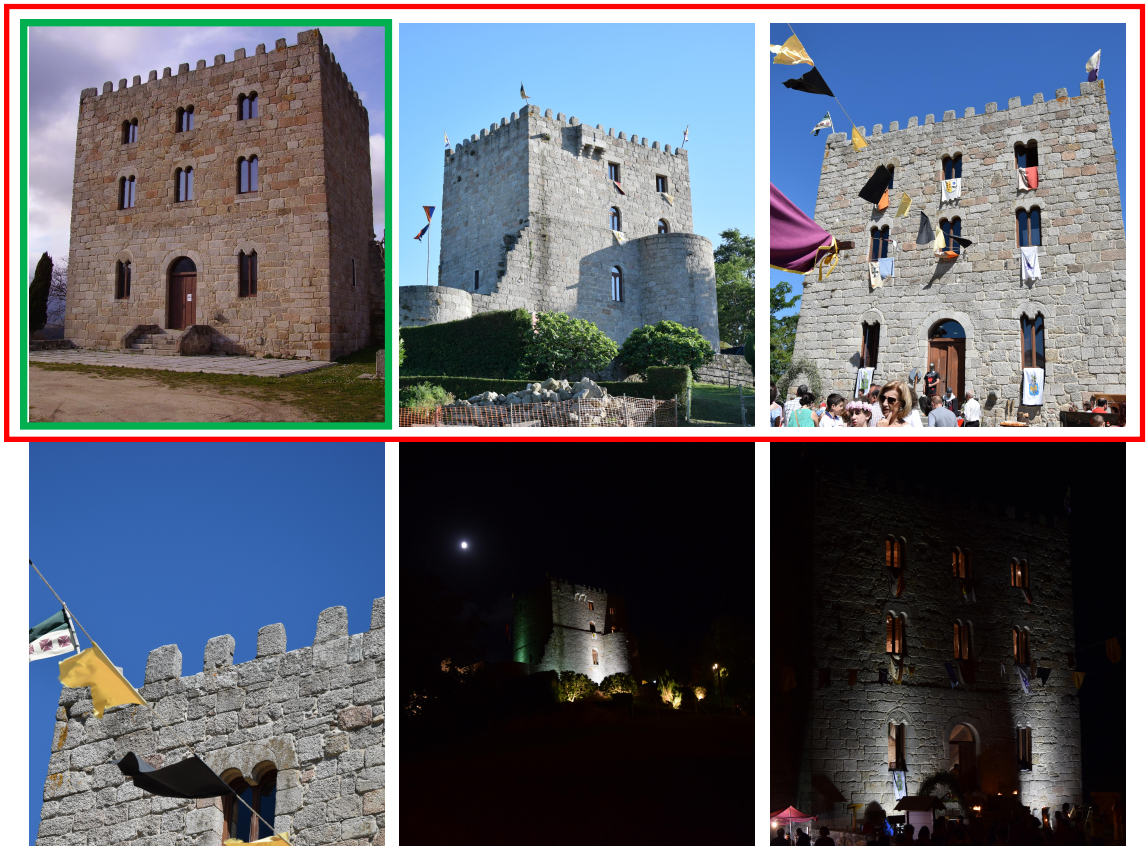}
    \caption{Example for the image selection methodology (the Castle of Pardo de Cela).
     \textbf{Red}: The top3 images with the highest similarity scores. 
     \textbf{Green}: The final image obtained through weighted sampling.}
    \label{fig:Effectiveness of the selection method}
\end{figure}

In the first stage (~\Cref{fig:pipeline}), we select representative images from the GLDv2 dataset. Since each landmark name corresponds to multiple images, typically sourced from Wikipedia entries related to the landmark, we use the landmark’s simple name (\eg “Eiffel Tower”) as the sole reference. Using CLIP-based similarity measures, we select the image that best matches this name, filtering out low-quality or ambiguous photos and ensuring a high-quality visual representation aligned with the landmark’s identity. From the top three images with the highest similarity scores, we conduct weighted sampling based on their similarity to select a unique image corresponding to each landmark.~\Cref{fig:Effectiveness of the selection method}
presents a specific example of image selection.
\begin{figure}[t!]
    \centering
    \includegraphics[width=\linewidth]{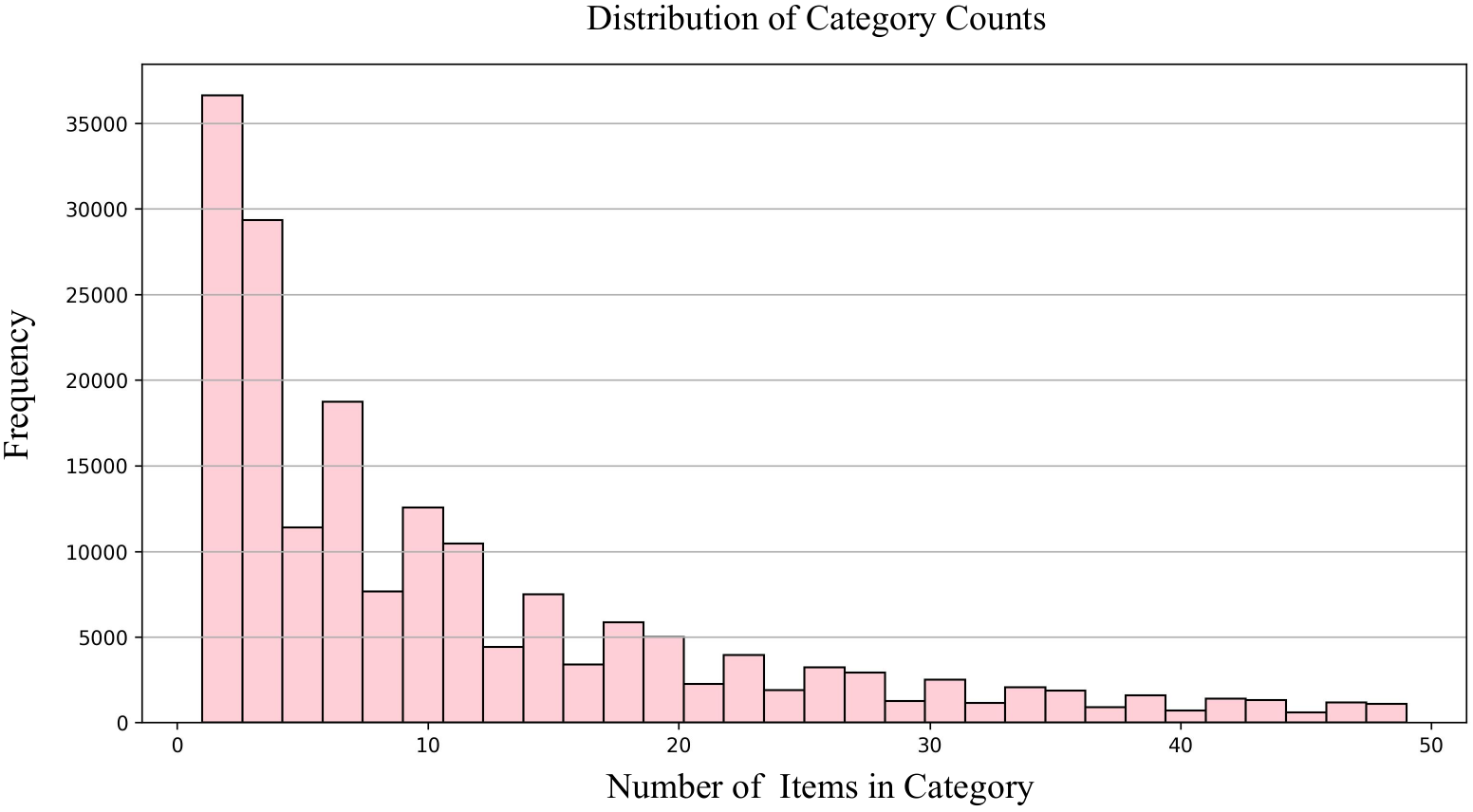}
    \caption{Frequency of the counts of images per category}
    \label{fig:distribution_of_category_count}
\end{figure}
\subsection{Prompt design for data annotation}\label{subappendix:prompt for annotation}
\begin{figure*}[htbp]
    \centering
    \includegraphics[width=\linewidth]{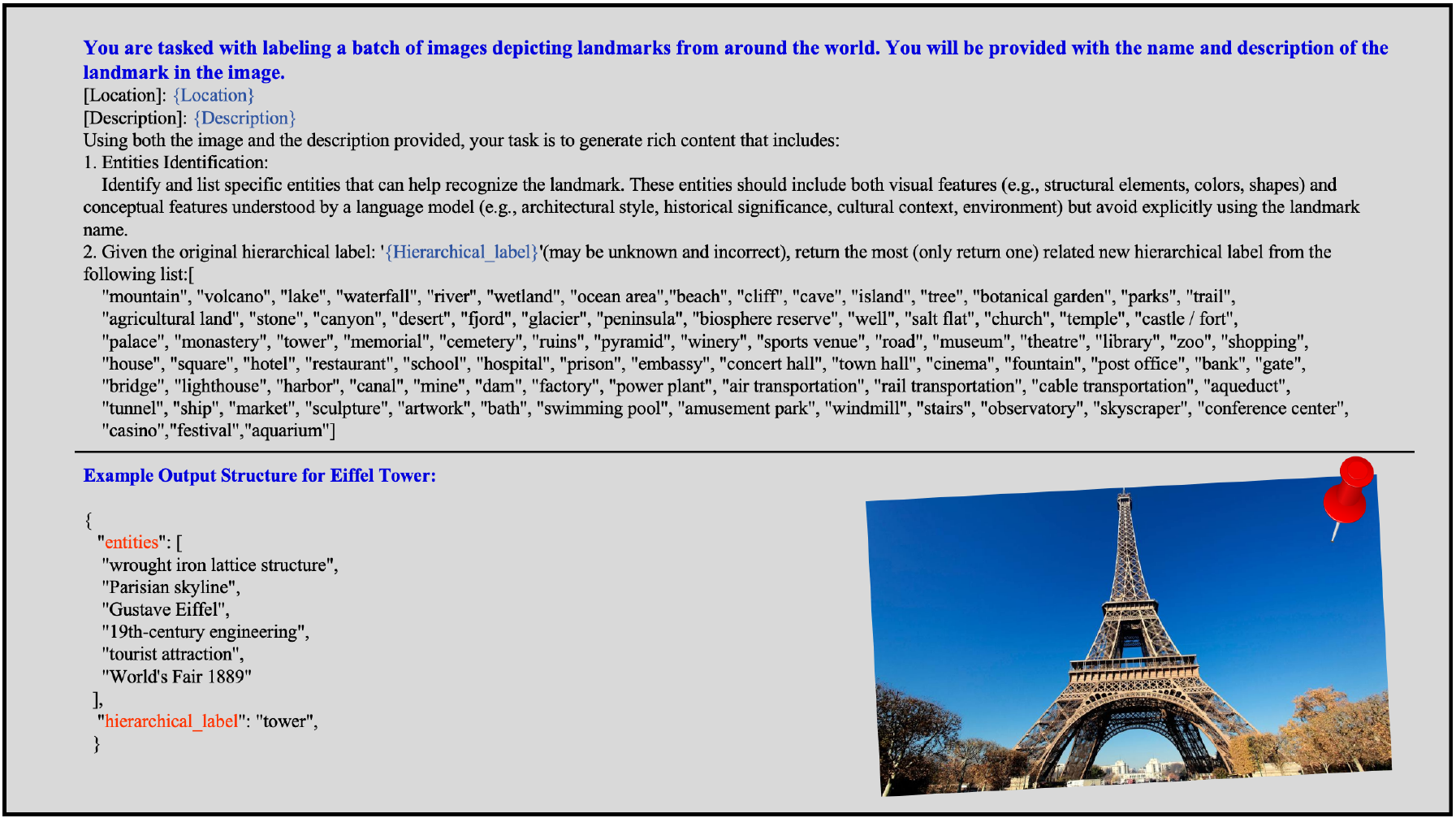}
    \caption{Multi-Granularity Data Generation Prompt. The description is the Q-A pair.}
    \label{fig:multi_granularity_data_generation_prompt}
\end{figure*}
\begin{figure}[t!]
    \centering
    \includegraphics[width=0.8\linewidth]{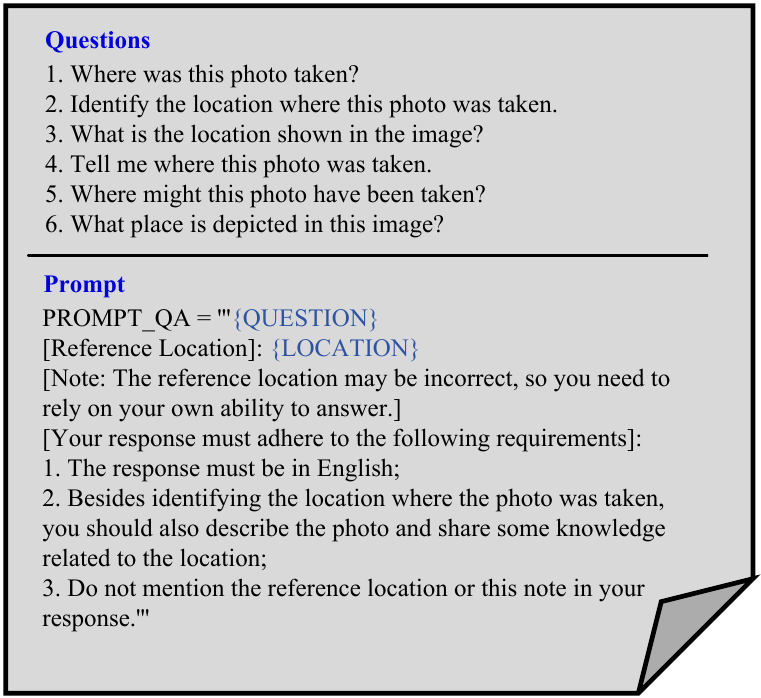}
    \caption{Q-A pair Prompt.}
    \label{fig:question_prompt}
\end{figure}

\paragraph{Q-A pair.}
Once we have a refined set of images, we generate Question-Answer (Q-A) pairs to facilitate landmark recognition. The prompt used for image Q-A pair annotation is shown in~\Cref{fig:question_prompt}. For each selected image, we random select one question from the questions set and add the landmark name as a reference to ensure the accuracy of the annotation. While answering these questions, the model is encouraged to provide descriptive details about the landmark, drawing on both the visual features and contextual information. This approach aims to broaden the model's understanding of each landmark’s visual and contextual uniqueness, laying a foundation for aligning with VE's cognitive framework. 

\paragraph{Multi-granularity data annotation.}
Following the Q-A pair annotation for each image, we generate multi-granularity data using the multi-granularity data generation prompt shown in~\Cref{fig:multi_granularity_data_generation_prompt}. In the original dataset, some hierarchical labels were already provided. To enhance accuracy, we employed GPT-4o to refine and expand the annotations, using the original labels as a reference. This process provides us with entities and an updated hierarchical label.
\subsection{MGLD overview.}
\begin{figure}[t!]
    \centering
    \includegraphics[width=\linewidth]{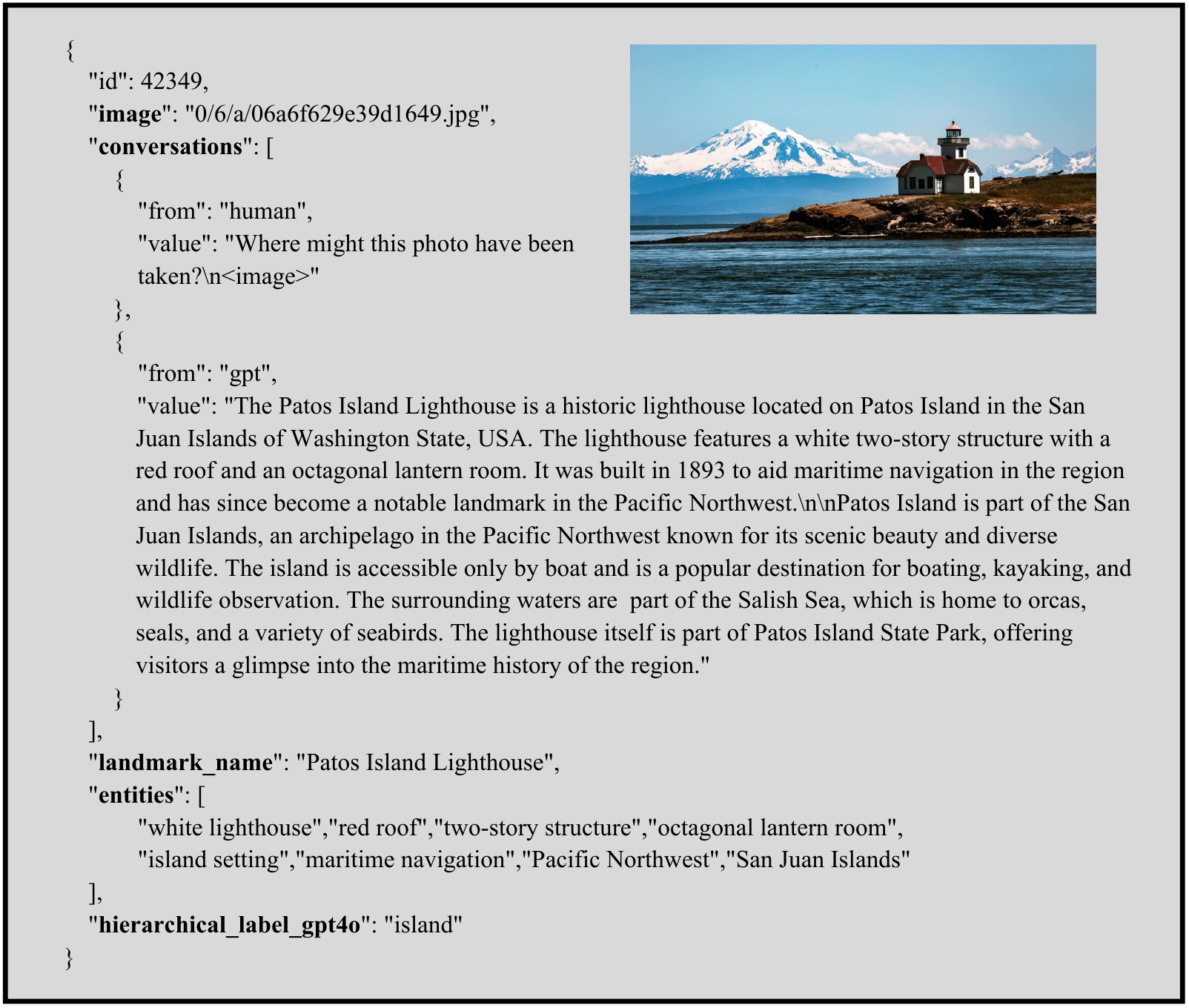}
    \caption{Example of the MGLD datasets. The conversation is the Q-A pair, and the hierarchical\_label\_gpt4o is the new hierarchical label annotated by GPT-4o. }
    \label{fig:multi_granularity_data_example}
\end{figure}

We structure the data as illustrated in~\Cref{fig:multi_granularity_data_example}. Each image is associated with a Q-A conversation, its landmark name, entities that capture both visual and conceptual features, and a hierarchical label representing its general category.

From the final dataset of approximately 203k samples, we set aside 5k samples as the test dataset.

\section{Evaluation Detail} 
\label{appendix: evaluation and statistics}

\subsection{Prompt design}
\begin{figure}[t!]
    \centering
    \includegraphics[width=0.8\linewidth]{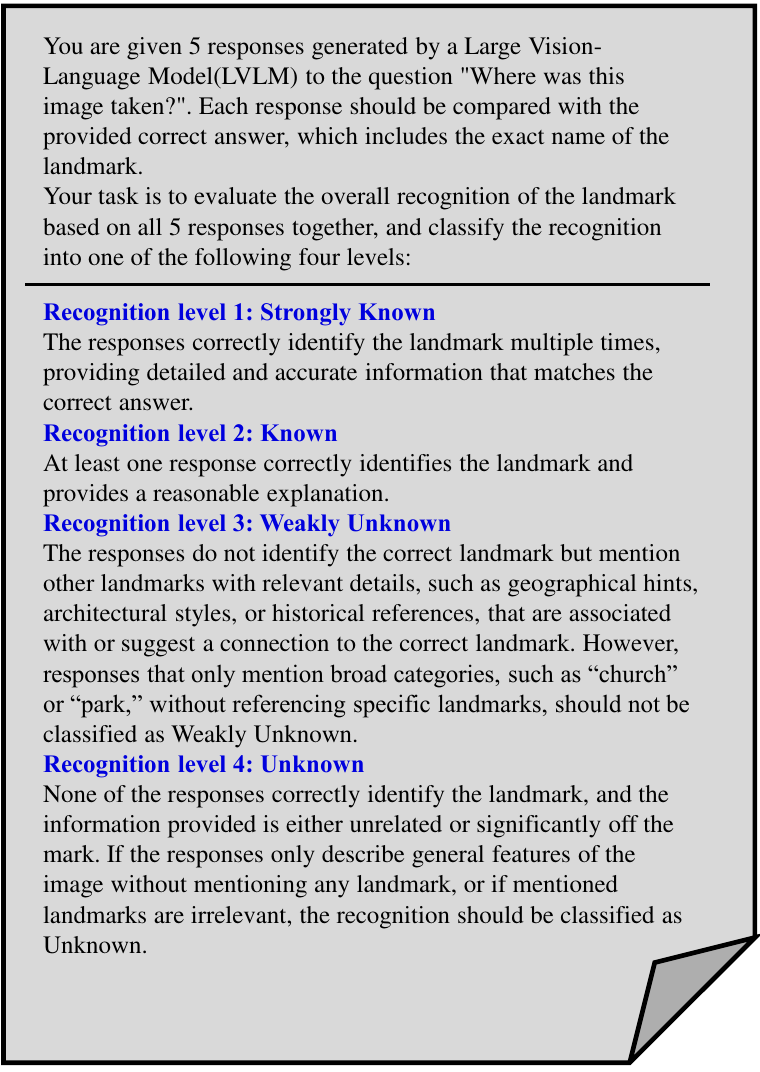}
    \caption{Evaluation Prompt}
    \label{fig:evaluation_prompt}
\end{figure}

We provide GPT4o with 5 inference runs of the VLLM and the ground-truth answer. 
The GPT4o is asked to evaluate the overall recognition of the landmark based on all 5 responses together, 
and classify the level of recognition into one of the four levels: Strongly Known, Known, Partially Known, or Unknown.
The classification criteria is clearly defined in the prompt (see~\Cref{fig:evaluation_prompt}).

\subsection{Rating criteria examples}\label{subappendix:detailed example}
\begin{figure}[t!]
    \centering
    \includegraphics[width=0.8\linewidth]{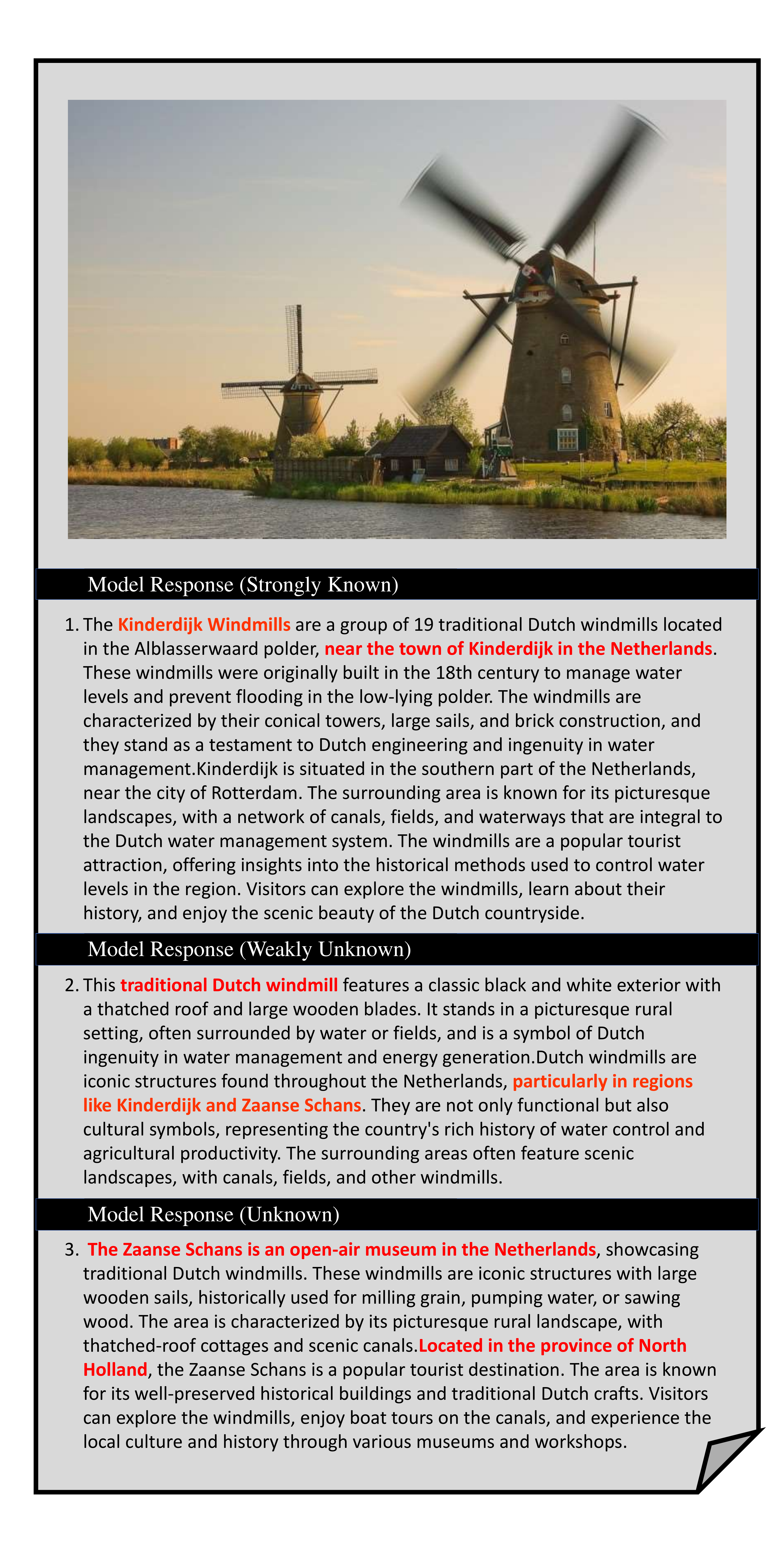}
    \caption{Model responses from different recognition level. There is a significant gap in recognition ability among ``Strongly Known," ``Weakly Unknown," and ``Unknown". The difference between ``Strongly Known" and ``Known" is the number of times the model correctly identifies the landmark.}
    \label{fig:evaluation_example}
\end{figure}

This section provides an example (``Kinderdijk Windmills``) evaluated by GPT-4o, where the answer across different models is assessed at four different levels—\textit{Strongly Known, Known, Partially Known, and Unknown}. ~\Cref{fig:evaluation_example} shows the specific answers from each model, followed by GPT-4o's explanations for the corresponding evaluations.

\begin{itemize}
    \item Explanation of strongly known : \textbf{The responses correctly identify the Kinderdijk Windmills multiple times}, specifically in Responses 2 and 3, providing detailed and accurate information that matches the correct answer. These responses include precise details about the location, historical significance, and characteristics of the Kinderdijk Windmills, demonstrating a strong recognition of the landmark.
    \item Explanation of known: Response 1 \textbf{correctly identifies the Kinderdijk Windmills} and provides detailed and accurate information about their location, history, and significance. This response alone is sufficient to classify the recognition as `Known'. Although Responses 2 and 3 incorrectly identify the location as Zaanse Schans, they do \textbf{provide relevant details about Dutch windmills}, which are related to the correct landmark. Responses 4 and 5 describe general features of Dutch windmills and mention regions like Kinderdijk, but do not specifically identify the landmark. Overall, the presence of one accurate identification with detailed information justifies the `Known' classification.
    \item Explanation of weakly unknown: None of the responses correctly \textbf{identify the `Kinderdijk Windmills' specifically}. However, several responses \textbf{mention relevant Dutch windmills and regions associated with windmills}, such as Kinderdijk and Zaanse Schans, providing geographical hints and cultural context related to Dutch windmills. This suggests a connection to the correct landmark, even though it is not explicitly named.
    \item Explanation of unknown: \textbf{None of the responses correctly identify the `Kinderdijk Windmills'}. The responses mention other windmills and locations such as `Weerschipmolen', `Zaanse Schans', and `Huis op de Hoop', but these are not related to the correct landmark. The information provided is either unrelated or significantly off the mark, as none of the responses provide any specific details or hints that connect to the Kinderdijk Windmills.
    
\end{itemize}

\section{Statistical Analysis for Partial Knowledge Parts}
\label{appendix:detailed accuracy}

\subsection{Detail results for ablation study}
We reports the results of the ablation study in terms of proportions relative to the entire dataset (see~\Cref{tab:ablation}). Here, in ~\Cref{tab:ablation_detail}, we provide the corresponding absolute counts for each knowledge level (Strongly Known, Known, Weakly Unknown, Unknown).
\begin{table*}[htbp]
\centering
\small
\setlength\tabcolsep{1pt} 
\begin{tabular}{lccccc}
\toprule
\textbf{Method} & \textbf{Strongly Known} & \textbf{Known} & \textbf{Weakly Unknown} & \textbf{Unknown} \\
\midrule
\textit{Baseline} &103&114&145&2138\\
\cellcolor{mygreen}
\textit{+ \textit{HSS-50k}} &187&161&145&2007\\
\cellcolor{mygreen}
\textit{+ \textit{HR-Branch}} &198	&149&152&2001\\
\cellcolor{mygreen}
\textit{+ $\mathcal{L}_e$} &212	&148&163&1977\\
\cellcolor{mygreen}
\textit{+ $\mathcal{L}_h$} &213&175&159&1953\\
\bottomrule
\end{tabular}
\caption{Detailed ablation study results. Absolute counts for each level of knowledge.}
\label{tab:ablation_detail}
\end{table*}
\subsection{Detailed results for generalizability experiment}
We summarizes the results of the generalizability experiments using accuracy (see~\Cref{tab:eeca_generalizability}). To complement this, ~\Cref{tab:generalizability_experiment_absolute} presents the absolute counts for each knowledge level (Strongly Known, Known, Weakly UnKnown, and Unknown).

\begin{table}[htbp]
\centering
\small
\setlength\tabcolsep{1pt} 
\begin{tabular}{lcccc}
\toprule
\textbf{Method} & \textbf{Strongly Known} & \textbf{Known} & \textbf{Weakly Unknown} & \textbf{Unknown} \\
 \midrule
\cellcolor{lightgray}Baseline & \cellcolor{lightgray}103& \cellcolor{lightgray}114& \cellcolor{lightgray}145& \cellcolor{lightgray}2138 \\
\midrule
\multicolumn{5}{c}{For HDS-25k} \\
\midrule 
 +Data & 182 & 142 & 151 & 2025 \\
 +HR Branch & 218 & 122 & 185 & 1975 \\
 +$\mathcal{L}_h$ & 233 & 127 & 175 & 1965 \\
 +$\mathcal{L}_e$ & 229 & 121 & 199 & 1951 \\
\midrule
\multicolumn{5}{c}{For HSS-25k} \\
\midrule
+Data & 169 & 135 & 201 & 1995 \\
 +HR Branch & 202 & 144 & 177 & 1977 \\
 +$\mathcal{L}_h$ & 205 & 141 & 193 & 1961 \\
 +$\mathcal{L}_e$ & 226 & 134 & 153 & 1987 \\
\midrule
\multicolumn{5}{c}{For LCS-25k} \\
\midrule
 +Data & 179 & 88 & 253 & 1980 \\
 +HR Branch & 203 & 99 & 243 & 1955 \\
 +$\mathcal{L}_h$ & 192 & 116 & 254 & 1938 \\
 +$\mathcal{L}_e$ & 198 & 110 & 266 & 1926 \\
\bottomrule
\end{tabular}
\caption{Generalizability experiment results. Absolute counts for each knowledge level across datasets and methods.}
\label{tab:generalizability_experiment_absolute}
\end{table}

\end{document}